\documentclass{ieeeaccess}
\usepackage{cite}
\usepackage{amsmath,amssymb,amsfonts}
\usepackage{algorithmic}
\usepackage{graphicx}
\usepackage{textcomp}
\def\BibTeX{{\rm B\kern-.05em{\sc i\kern-.025em b}\kern-.08em
    T\kern-.1667em\lower.7ex\hbox{E}\kern-.125emX}}

\usepackage{multirow}
\usepackage{tabularx}
\usepackage{booktabs}
\usepackage[switch]{lineno}  
\usepackage{rotating}
\usepackage{siunitx}
\usepackage[labelformat=simple]{subcaption}

\usepackage{comment}
\usepackage{enumitem}
\usepackage{hyperref}
\usepackage[table,xcdraw]{xcolor}

\def\ie{\emph{i.e.}}
\def\etal{\emph{et al.}}

\def\T{\mathcal{T}}
\DeclareMathOperator{\spn}{span}

\begin{document}

\history{Date of publication March 29, 2023, date of current version April 4, 2023.}
\doi{DOI 10.1109/ACCESS.2023.3263210}

\title{Probing the Purview of Neural Networks via Gradient Analysis}
\author{
\uppercase{Jinsol Lee, \IEEEmembership{Member, IEEE}, Charlie Lehman, Mohit Prabhushankar, \IEEEmembership{Member, IEEE}}, 
\uppercase{and Ghassan AlRegib}, \IEEEmembership{Fellow, IEEE}}
\address{Omni Lab for Intelligent Visual Engineering and Science (OLIVES)\\ School of Electrical and Computer Engineering, Georgia Institute of Technology, Atlanta, GA 30332, USA}

\markboth
{J. Lee \headeretal: Probing the Purview of Neural Networks via Gradient Analysis}
{J. Lee \headeretal: Probing the Purview of Neural Networks via Gradient Analysis}

\corresp{Corresponding author: Jinsol Lee (e-mail: jinsol.lee@gatech.edu).}

\begin{abstract}
We analyze the data-dependent capacity of neural networks and assess anomalies in inputs from the perspective of networks during inference.
The notion of data-dependent capacity allows for analyzing the knowledge base of a model populated by learned features from training data.
We define \textit{purview} as the additional capacity necessary to characterize inference samples that differ from the training data.
To probe the \textit{purview} of a network, we utilize gradients to measure the amount of change required for the model to characterize the given inputs more accurately.
To eliminate the dependency on ground-truth labels in generating gradients, we introduce \textit{confounding labels} that are formulated by combining multiple categorical labels.
We demonstrate that our gradient-based approach can effectively differentiate inputs that cannot be accurately represented with learned features.
We utilize our approach in applications of detecting anomalous inputs, including out-of-distribution, adversarial, and corrupted samples.
Our approach requires no hyperparameter tuning or additional data processing and outperforms state-of-the-art methods by up to 2.7\%, 19.8\%, and 35.6\% of AUROC scores, respectively.
\end{abstract}

\begin{keywords}
Gradients, model capacity, robustness, out-of-distribution detection, adversarial detection
\end{keywords}

\titlepgskip=-15pt

\thispagestyle{empty}
\twocolumn[{%
\vspace{30mm}
{ \large
\begin{itemize}[leftmargin=2.5cm, rightmargin=0.8cm, align=parleft, labelsep=2.2cm, itemsep=4ex,]

\item[\textbf{Citation}]{J. Lee, C. Lehman, M. Prabhushankar and G. AlRegib, “Probing the Purview of Neural Networks via Gradient Analysis,” in \textit{IEEE Access}, vol. 11, pp. 32716-32732, 2023, doi: 10.1109/ACCESS.2023.3263210.}

\item[\textbf{BibTeX}] {
  @ARTICLE\{lee2023purview,\\
  author=\{Lee, Jinsol and Lehman, Charlie and Prabhushankar, Mohit and AlRegib, Ghassan\},\\
  journal=\{IEEE Access\}, \\
  title=\{Probing the Purview of Neural Networks via Gradient Analysis\}, \\
  year=\{2023\},\\
  volume=\{11\},\\
  pages=\{32716-32732\},\\
  publisher=\{IEEE\},\\
  doi=\{10.1109/ACCESS.2023.3263210\}\}}

\item[\textbf{Copyright}]{\textcopyright2023 Lee, Lehman, Prabhushankar and AlRegib. This is an open-access article distributed under the terms of the Creative Commons Attribution License (CC BY). The use, distribution or reproduction in other forums is permitted, provided the original author(s) and the copyright owner(s) are credited and that the original publication in this journal is cited, in accordance with accepted academic practice. No use, distribution or reproduction is permitted which does not comply with these terms.}

\item[\textbf{Contact}]{
\{jinsol.lee, alregib\}@gatech.edu\\
\url{https://ghassanalregib.info/}\\}
\end{itemize}
}}]
\newpage
\clearpage

\setcounter{page}{1}

\maketitle

%%%%%%%%%%%%%%%%%%%%%%%%%%%%%%%%%%%%%%%%%%%%%%%%%%%%%%%%%%%%%%%
\section{Introduction}\label{sec:introduction}

Deep neural networks are prone to failure when deployed in real-world environments as they often encounter data that diverge from training conditions~\cite{temel2018cureor, hendrycks2019robustness, temel2019multifarious}. 
Neural networks rely on the implicit assumption that any given input during inference is drawn from the same distribution as the training data. 
Limited to classes seen during training, neural networks classify any input image among such in-distribution classes, even if the image is significantly different from training data. 
In addition, it is widely accepted that neural networks tend to make overconfident predictions even for inputs that differ from training data~\cite{goodfellow2014adversarial, guo2017calibration, lee2020gradients, lee2022gradient}, making it more challenging to distinguish inputs of anomalous conditions. 
This behavior can have serious consequences when utilized in safety-critical applications, such as autonomous vehicles and medical diagnostics~\cite{temel2017curetsr,Logan2022ISBI}. 
To ensure reliable performance for practical applications of neural networks, models must be able to distinguish inputs that differ from training data and cannot be handled adequately based on their capacity.

The capacity of neural networks is broadly discussed in terms of the size of the networks (\ie, the number of model parameters)~\cite{bengio2009learning, dauphin2013big, goodfellow2016deep}. 
It is central to the generalization performance of neural networks in traditional statistical learning theories as models with larger capacity are expected to overfit training data, leading to poor generalization performance~\cite{vapnik1971uniform, goodfellow2016deep}.
However, recent studies show that over-parameterization of deep neural networks only helps with their generalization performance~\cite{neyshabur2014search, zhang2021understanding}.
Many researchers aim to understand this phenomenon by examining the \textit{representational capacity}, which is the types of functions a model \textit{can} learn~\cite{cybenko1989approximation, montufar2014number}. 
Others discuss it from the optimization point of view~\cite{goodfellow2016deep}.
They recognize that a learning algorithm, defined by model architecture and training procedure, is unlikely to find the best function among all possible functions, although it can still learn one that performs well for a given task.  
This notion of capacity, defined as the types of functions that can be reached via some learning algorithm, is denoted as the \textit{effective capacity}~\cite{neyshabur2014search, zhang2021understanding}.
These perspectives of data-independent network capacity analyze the generalization behavior of neural networks by changing network architectures or forcing memorization.
They assume that training data is available to be manipulated for memorization and repetitive training of networks with altered architectures.
However, this assumption does not hold for deployed models in which the training data and network architectures are fixed and novel situations cannot be foreseen or simulated during training.
Hence, they are not suitable for analyzing the capacity of deployed models and their ability to handle samples given during inference.

\begin{figure}[t]
    \centering
    \includegraphics[width=0.99\linewidth]{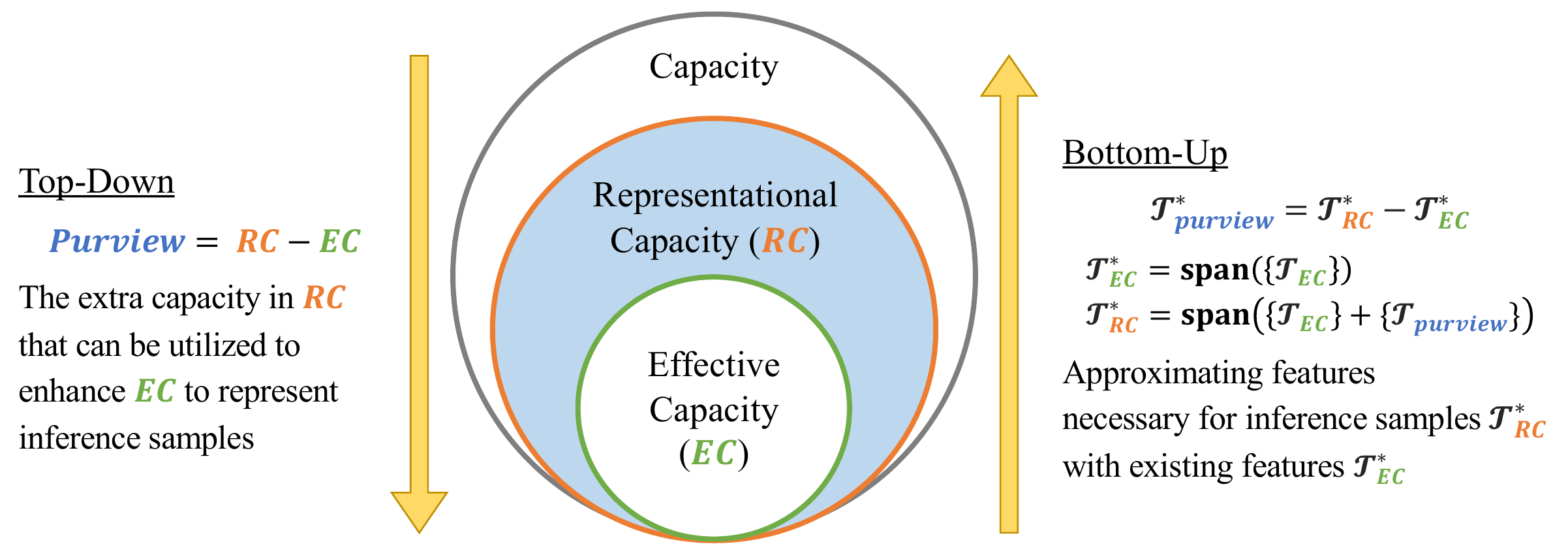}
    \caption{Definition of \textit{purview}, shown in the region highlighted in blue. A top-down \textit{purview} is established on the inclusive relationship between $RC$ and $EC$. A bottom-up \textit{purview} depends on learned features.}
    \label{fig:purview}
\end{figure}

In this work, we examine the capacity of trained neural networks in terms of their knowledge base and reformulate the definition of the model capacity.
The knowledge base of a network is established by training data, allowing the model to characterize the inputs observed during inference with its learned features.
We argue that the capacity of models should be investigated not only in terms of the learning algorithm (\ie, model architectures and training procedures) but also in terms of the training data. 
To analyze the data-dependent capacity of the trained networks, we introduce the concept of \textit{purview} from top-down and bottom-up perspectives, as illustrated in Fig.~\ref{fig:purview}.
The top-down \textit{purview}, derived from the model perspective, is built on the inclusive relationship between \textit{representational capacity} ($RC$) and \textit{effective capacity} ($EC$), where the latter is a subset of the former.
The \textit{purview} lies in the region described by $RC - EC$ and shown in blue, which denotes the additional capacity in $RC$ that can be utilized to enhance $EC$.
The bottom-up \textit{purview}, established from the data perspective, is based on the learned features $\T$ that populate the knowledge base of a trained model, \ie, the data-dependent $EC$ and its features $\T_{EC}$.
It is the gap between the learned features $\T^{*}_{EC}$ established on the training data and the ideal $EC$ with additional features in $\T^{*}_{purview}$ that is necessary to handle inference samples in addition to the training data.

To probe the \textit{purview} of the trained networks, we utilize gradients to determine anomalous inputs from the perspective of the models.
Our intuition is that gradients correspond to the amount of change that a model requires to properly represent a given sample.
We hypothesize that the required change captured in gradients would be more significant for inputs that cannot be represented accurately with the learned features of a network.
From the data-dependent view of model capacity, a model trained with simpler data would have a less extensive feature set, and thus, lower $EC$.
In turn, there is more room for improvement in its feature set (\ie, larger $RC - EC$ region, larger \textit{purview}), leading to more significant gradient responses.
During training, gradients are generated with respect to the model outputs and ground-truth labels for given inputs, the latter of which are unavailable during inference.
To remove dependency on information regarding given samples during inference, we introduce \textit{confounding labels}, which are labels formulated by combining multiple categorical labels.
Although the use of manipulated labels has been explored in existing studies~\cite{tokozume2018between, zhang2017mixup, yun2019cutmix, durand2019learning, duarte2021plm}, it is only analyzed during model training, and the label designs rely on the statistics of the training data.
In comparison, a \textit{confounding label} does not require any knowledge of the training data and can be used during inference with no access to ground-truth labels.
The contributions of this paper are five-fold:

\begin{itemize}[topsep=2mm,leftmargin=4mm,itemsep=1mm]
    \item We define \textit{purview} by integrating the top-down view from the \textit{representational capacity} to the \textit{effective capacity} and the bottom-up view based on the learned features of the trained models.
    \item We introduce the concept of \textit{confounding labels} as an unsupervised tool to elicit a model response that can be utilized to probe the \textit{purview} of trained neural networks.
    \item We demonstrate the relationship between the degree of exposure to diverse data in model training and the \textit{purview} of trained networks via gradient analysis.
    \item We utilize our proposed method in applications of OOD detection, adversarial detection, and corrupted input detection and achieve state-of-the-art performance.
    \item We conduct extensive ablation studies to study the manipulation of supervised and unsupervised \textit{confounding labels}.
\end{itemize}

%%%%%%%%%%%%%%%%%%%%%%%%%%%%%%%%%%%%%%%%%%%%%%%%%%%%%%%%%%%%%%%
%%%%%%%%%%%%%%%%%%%%%%%%%%%%%%%%%%%%%%%%%%%%%%%%%%%%%%%%%%%%%%%
\section{Related Work}\label{sec:literature}

%%%%%%%%%%%%%%%%%%%%%%%%%%%%%%%%%%%%%%%%%%%%%%%%%%%%%%%%%%%%%%%
\subsection{Capacity of Neural Networks}

Following the distinction between \textit{representational capacity} and \textit{effective capacity} established in Section~\ref{sec:introduction}, we introduce how they have been studied in the literature.
VC (Vapnik-Chervonenkis) dimension~\cite{vapnik1971uniform} is a traditional approach for quantifying the \textit{representational capacity} of a learning machine with cardinality of data samples to provide the theoretical bounds.
More recent studies have examined the \textit{representational capacity} of deep neural networks from the viewpoint of generalization, which diverges from the traditional understanding of the overfitting nature of larger models.
To do so, some proposed to utilize sample complexity to understand standard generalization~\cite{neyshabur2014search, mei2022towards} and adversarial generalization~\cite{schmidt2018adversarially}.
Others employed the intrinsic dimensions of networks~\cite{li_id_2018_ICLR, ansuini2019intrinsic}. 
Zhang~\etal~\cite{zhang2021understanding} proposed the concept of the \textit{effective capacity} to analyze the memorization behavior of deep neural networks, extended by Arpit~\etal~\cite{arpit2017closer}.
They demonstrated the memorization behavior by searching for the network of the smallest capacity required to learn perturbed images and labels.
These studies of \textit{representational capacity} and \textit{effective capacity} analyze networks during training by altering the model architectures.
Their interpretation of the model capacity is data-independent.
They focus on the varying sizes of the models and the functions they are capable of learning, rather than on the effect of the training data, in addition to the network architectures and training algorithms.
We argue that training data is critical in defining the \textit{effective capacity} of networks.
This data-dependent view of the \textit{effective capacity} of trained models allows for analyzing their ability to handle different inputs during inference and determining the validity of model predictions.
We aim to demonstrate the connection between \textit{representational capacity} and \textit{effective capacity} from a data-dependent perspective and the proposed concept of \textit{purview}.

%%%%%%%%%%%%%%%%%%%%%%%%%%%%%%%%%%%%%%%%%%%%%%%%%%%%%%%%%%%%%%%
\subsection{Anomalous Input Detection}

Many studies have aimed to detect anomalous samples due to statistical shifts in the data distribution or adversarial generation.
This application exploits the insufficient knowledge base of models trained on relatively small datasets to distinguish inputs that differ from the training samples.
There are mainly two types of approaches: analyzing the output distributions of classifiers to differentiate anomalous inputs or employing auxiliary networks for detection.
Hendrycks and Gimpel~\cite{hendrycks2016baseline} introduced a baseline method of thresholding samples based on predicted softmax distributions.
Liang~\etal~\cite{liang2018odin} employed additional input and output processing to the previous method to further improve detection performance.
DeVries and Taylor~\cite{devries2018confidence} utilized prediction confidence scores obtained from an augmented confidence estimation branch on a pretrained classifier. 
Ma~\etal~\cite{ma2018lid} proposed to characterize the dimensional properties of adversarial regions with local intrinsic dimensionality. 
Lee~\etal~\cite{lee2018mahalanobis} proposed a confidence metric using Mahalanobis distance. 
Liu~\etal~\cite{liu2020energyOOD} utilized energy scores to capture the likelihood of occurrence during inference and training time. 
These approaches utilize activation-based representations to characterize the anomaly in inputs via learned features that establish the \textit{effective capacity} of a model, \ie, what the network knows about the given inputs.
However, the overconfident nature of neural networks makes it counterintuitive to characterize anomalous inputs solely with activations that are known to be poorly calibrated~\cite{guo2017calibration}.
Rather, we argue that the anomaly in inputs should be established based on what the model is unfamiliar with and thus incapable of representing accurately.
Instead of focusing solely on the learned features based on the \textit{effective capacity} of the networks, we focus on the additional features necessary to represent the anomaly in the regions of \textit{purview}.

%%%%%%%%%%%%%%%%%%%%%%%%%%%%%%%%%%%%%%%%%%%%%%%%%%%%%%%%%%%%%%%
\subsection{Gradients as Features}\label{ssec:gradient_features}

At the core of the advancement of deep neural networks lies gradient-based optimization techniques that allow for finding solutions to tasks at hand~\cite{rumelhart1986learning}.
In addition to its utility as an optimization tool, gradients have been utilized for various purposes.
Goodfellow~\etal~\cite{goodfellow2014adversarial} first demonstrated that small, hardly perceptible perturbations, known as adversarial attacks, can deliberately fool trained networks to make irrelevant predictions with high confidence, extended by numerous studies including~\cite{kurakin2017bim_iterll, madry2017pgd}.
These gradient-based adversarial attacks force models to deviate from their data-dependent $EC$ and yield perturbed outputs.
Gradients are also commonly used in visual explanation techniques to produce localization maps of pixels relevant to model predictions~\cite{selvaraju2017gradcam, chattopadhay2018grad}, which accentuate the existing features that define the \textit{effective capacity}.
Others expanded their work with contrastive explanations, in which they visualized features of other potential predictions for actual predictions~\cite{alregib2022explanatory, prabhushankar2022introspective}.
Another application using gradients is anomaly detection by constraining gradients in autoencoder learning such that inliers form a specific knowledge base that can be used to differentiate outliers~\cite{kwon2020backpropagated}.
Some have utilized gradients to quantify the uncertainty of trained neural networks~\cite{oberdiek2018classification, lee2020gradients, huang2021gradnorm}.
In general, gradients are used to illuminate the learned features of networks and examine the divergence from the established knowledge base.
Our focus of \textit{purview} exploits this nature of gradients in terms of model capacity, bridging the gap between \textit{effective capacity} and \textit{representational capacity}.
We show that gradients enable the analysis of data samples during inference from the perspective of models by considering the adequacy of the \textit{effective capacity} established on training data and the additional capacity (\ie, \textit{purview}), stepping further into the \textit{representational capacity} needed for more accurate representation.

%%%%%%%%%%%%%%%%%%%%%%%%%%%%%%%%%%%%%%%%%%%%%%%%%%%%%%%%%%%%%%%
\subsection{Manipulating Label Encodings}

One-hot encoding is the most widely used approach for formulating labels to handle nominal data (\ie, categorical data with no quantitative relationship between categories).
A combination of multiple one-hot encodings is used to indicate information regarding multiple classes in each image. 
Based on these two, recent studies have analyzed different formulations of labels for various purposes.
Some work~\cite{tokozume2018between, zhang2017mixup, yun2019cutmix} have proposed 
the mixing of two input images and their labels with a pre-computed ratio as a data augmentation technique.
For the interpretability of neural networks, Prabhushankar and AlRegib~\cite{prabhushankar2021extracting} proposed to extract causal visual features using combinations of binary classification labels.
To address the problem of missing labels in a multi-label classification setting, Durand~\etal~\cite{durand2019learning} utilized partial labels formulated using the proportions of known labels.
Duarte~\etal~\cite{duarte2021plm} proposed a similar approach to handle imbalanced datasets by masking parts of the labels based on the ratio of positive samples to negative samples.
These methods impose constraints on feature learning to improve the generalizability and robustness.
However, they explored different label encodings only for model training and assumed the availability of information regarding the training data.
We propose manipulating label encodings to probe the \textit{purview} of networks built upon their knowledge base of learned features.
Contrary to existing studies, our approach does not depend on the availability of training data statistics or label information for the inference data.

\section{Probing the Purview of Neural Networks}

In this section, we expand the definition of the \textit{purview} of neural networks based on data-dependent \textit{effective capacity}.
We discuss the \textit{purview} of neural networks in terms of the knowledge base established by the training data and the additional knowledge necessary to handle samples during inference.
We then introduce our gradient-based approach to probe the \textit{purview} of the trained models.

%%%%%%%%%%%%%%%%%%%%%%%%%%%%%%%%%%%%%%%%%%%%%%%%%%%%%%%%%%%%%%%%
\subsection{Feature-Based Capacity Analysis} \label{ssec:capacity}

We define \textit{purview} as the additional capacity required for a model to handle samples observed during inference that differ from the training data.
Our definition of \textit{purview} depends on the \textit{effective capacity} established by the training data, as well as network architectures and training procedures.
Assuming a fixed architecture and training procedure, we first discuss the \textit{effective capacity} of a trained model in terms of the features learned from the training data.
Following the notations utilized by AlRegib and Prabhushankar~\cite{alregib2022explanatory}, we introduce our setup for feature-based \textit{effective capacity} analysis. % and connect it to \textit{purview}.
Let $f(\cdot)$ be a neural network trained for an image classification task,
\begin{equation}
    f : \mathbb{R}^{h \times w \times c} \rightarrow \mathbb{R}^N,
\end{equation}
where it maps an input of dimension $h \times w \times c$ to an output vector of dimension $N$, which is the number of classes defined in the training dataset.
Given an input image $x$, the model $f$ utilizes its learned features to produce a model output (\ie, logits)
\begin{equation}
    y = f(x).
\end{equation}
The predicted class $C$ is determined by taking the index of the largest logit value,
\begin{equation}
    C = \operatorname*{arg\,max}~y, ~~C \in \{1, 2, \cdots, N\}. 
\end{equation}
Let $\T$ be the set of all features that the network learned to extract during training for the classification task,
\begin{equation}
    \T = \left\{ \T_1, \T_2, \cdots, \T_P \right\},
\end{equation}
where $P$ denotes the total number of learned features.
Each feature captures a unique characteristic of the training data.
Depending on the classes represented in the training data and their similarities, some classes have unique features whereas others have shared features for their similar characteristics.
The model prediction of $C$ for image $x$ means that the network's decision is based on the set of features relevant to class $C$, which is a subset of all available features $\T$.

\begin{figure*}[ht]
    \centering 
    \begin{subfigure}[t]{.38\textwidth}
        \includegraphics[width=\textwidth]{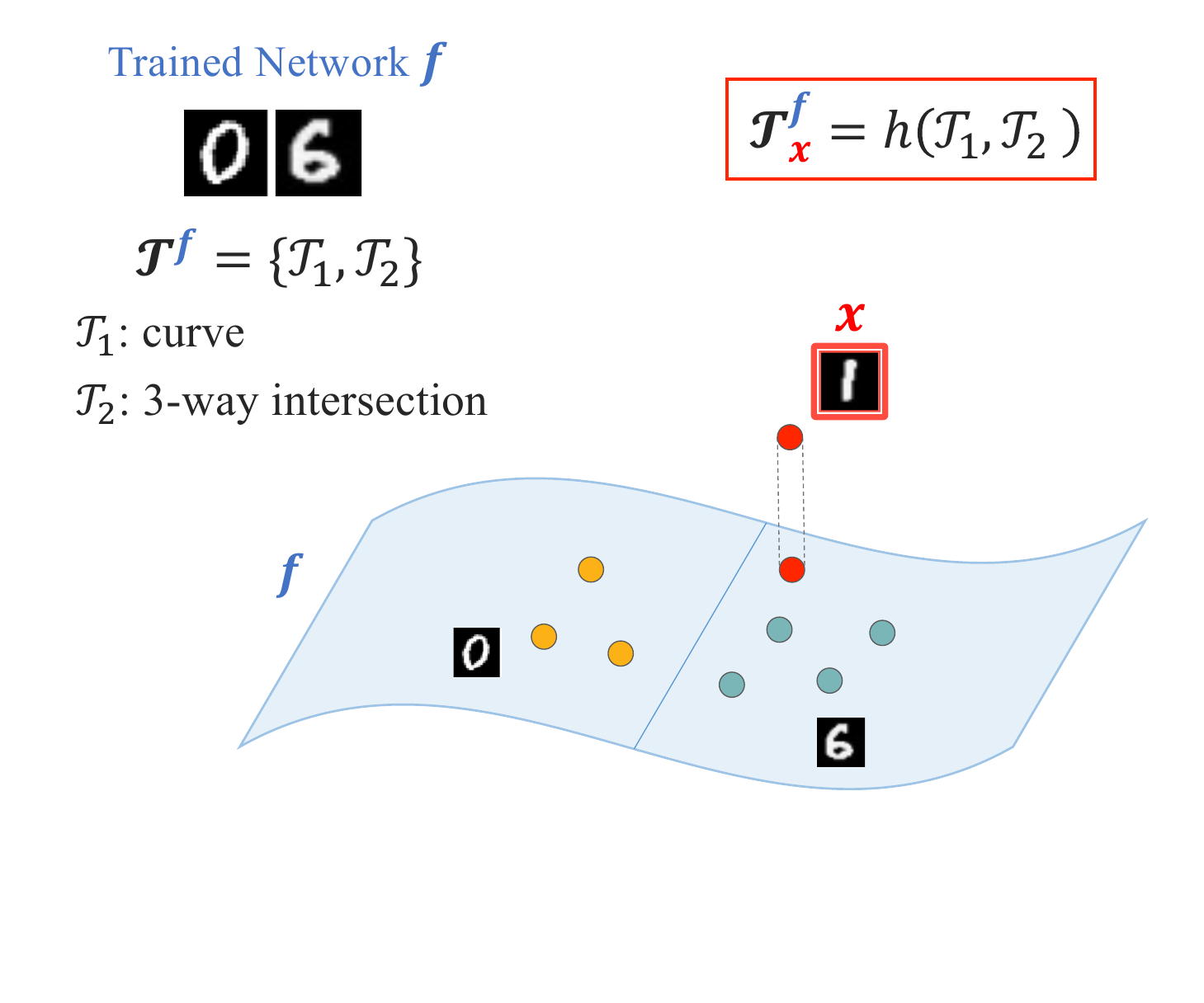}
        \caption{Model $f$ approximating for $x$}
        \label{sfig:manifold_f}
    \end{subfigure}
    \hspace{1.5mm}\hspace{1.5mm}
    \begin{subfigure}[t]{.38\textwidth}
        \includegraphics[width=\textwidth]{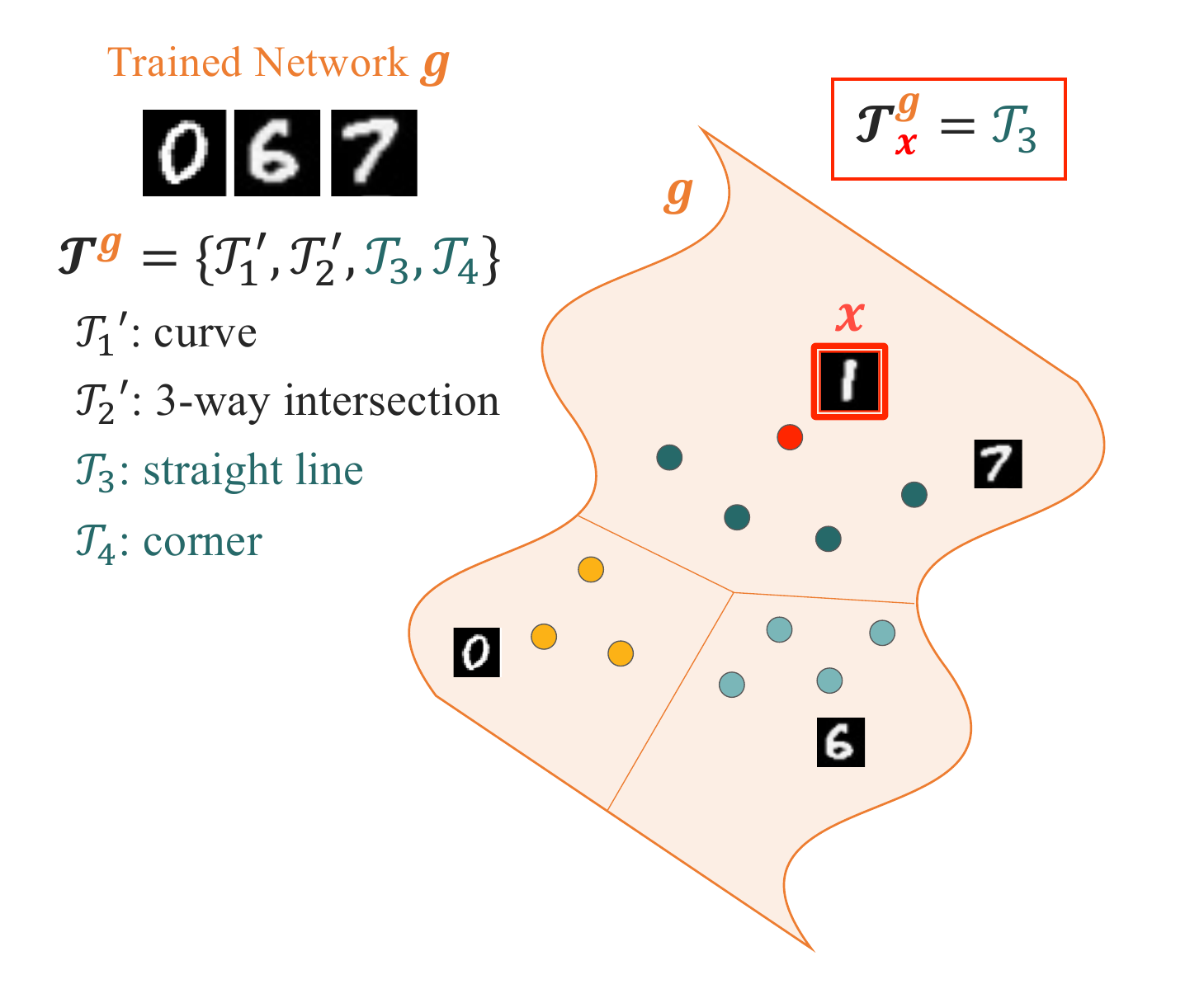}
        \caption{Model $g$ using an appropriate feature for $x$}
        \label{sfig:manifold_g}
    \end{subfigure}
    \hspace{1mm}
    \begin{subfigure}[t]{.17\textwidth}
        \includegraphics[width=\textwidth]{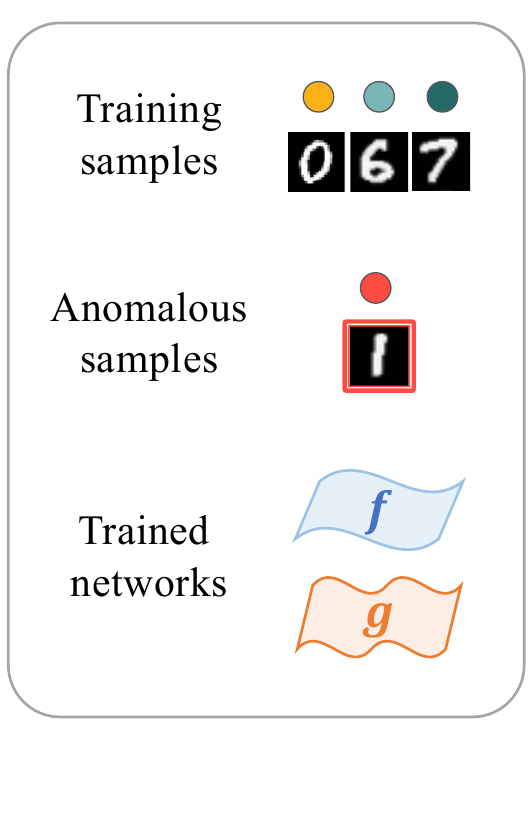}
        \setcounter{subfigure}{0}
        \label{sfig:manifold_legend}
    \end{subfigure}
\vspace{1mm}
\caption{Illustrations of the feature-based \textit{effective capacity} of trained neural networks established by their training data, given that the networks share the same architecture and training procedures.}
\label{fig:manifolds}
\end{figure*}\vspace{1mm}

We describe our interpretation of data-dependent \textit{effective capacity} using the toy example in Fig.~\ref{fig:manifolds}.
Consider two networks, $f$ and $g$, which share the same architecture and training procedures, but have different training data.
We assume that the network architecture has a sufficient number of model parameters to learn their training data and that the training procedures allow for optimization.
The first network $f$ is trained on handwritten digit images of $0, 6$, shown in Fig.~\ref{sfig:manifold_f}.
Based on the training data, it learns the features of curves shared among all digits ($\T_1$) and 3-way intersections for the digit $6$ ($\T_2$), forming the feature set $\T^f$,
\begin{equation}
    \T^f = \{ \T_1, \T_2 \}.
\end{equation}
On the other hand, the second network $g$ is trained on the images of the same handwritten digits as $f$ and an additional digit of $7$. 
Some of the learned features of $g$ are similar to those of $f$, and are written as $\T'_1$ and $\T'_2$.
Moreover, the additional training data of digit $7$ leads to extra features of straight lines ($\T_3$) and corners ($\T_4$), constructing the overall feature set $\T^g$,
\begin{equation}
    \T^g = \{ \T_1', \T_2', \T_3, \T_4 \}.
\end{equation}
Model $g$ has a higher \textit{effective capacity} with a more comprehensive feature set due to its exposure to relatively diverse training data compared to $f$, which is described with more ridges in Fig.~\ref{sfig:manifold_g}.
Given these two models, we now consider an inference scenario in which an input image $x$ of the handwritten digit $1$ is presented.
For model $f$, the feature span $\T^f$ is insufficient to accurately represent $x$ due to the lack of the features for the straight lines.
The best the model $f$ can do to capture the characteristics of the sample $x$ is to utilize the existing features for an approximation,
\begin{equation}
    \T^f_x = h(\T_1, \T_2),
\end{equation}
where $h$ is a function that combines the learned features.
For model $g$, however, the same image of digit $1$ can be handled more properly because the feature set $\T^g$ includes the straight-line feature, learned from digit $7$ during training.
The model $g$ can utilize this learned feature to represent the inference sample, 
\begin{equation}
    \T^g_x = \T_3.
\end{equation}
In contrast to $\T^g_x$, $\T^f_x$ is still inadequate for representing the sample because each learned feature captures a unique characteristic, and the combination of existing features cannot precisely account for the lack of relevant feature.
While the predictions from both models are irrelevant because the class of $1$ does not exist in the training data, this feature-based focus enables the analysis of the anomaly in inputs from the perspective of models that is not limited to the predicted class distributions.

The comparison between these two models highlights the core of our approach, which focuses on the absence of relevant features in the model's knowledge base to distinguish inputs that cannot be represented accurately, \ie, anomalous samples.
Although the knowledge base of a model can be broadened via training with comprehensive datasets and various data augmentation techniques, it is still impractical for a model to learn everything in existence.
Consequently, in practice, we only have access to models that lack some features (\ie, $f$).
Writing in general terms, 
\begin{align}
    \T^f &= \{ \T_1, \cdots, \T_P \}, \\
    \T^g &= \{ \T_{P+1}, \cdots, \T_{P+Q} \}
\end{align}
Ideally, we want the approximation $\T^f_x$ to be equivalent to the lacking feature, 
\begin{equation}
    h(\T_1, \cdots, \T_P) = \T_{P+1},
\end{equation}
where $\T_{P+1}$ is an arbitrary feature that is relevant to input $x$ and exists only in $\T^g$.
This would allow for the representation of the sample considered anomalous from the perspective of the models with its learned features.
With the \textit{purview}, we examine the gap in the knowledge base of $f$ that needs to be offset to fully grasp the unfamiliar characteristics of $x$.

%%%%%%%%%%%%%%%%%%%%%%%%%%%%%%%%%%%%%%%%%%%%%%%%%%%%%%%%%%%%%%%%
\subsection{Purview of Networks and Gradients}\label{ssec:purview}

Building on the \textit{effective capacity} analysis setup, we discuss the \textit{purview} of the models.
We define \textit{purview} as the additional feature-based capacity necessary for a trained network to characterize inputs given during inference that differ from the training data.
It is centered on the absence of features in the network's feature span that are relevant for the given inputs.
Consider the models $f$ and $g$ and their features used in response to input $x$ during inference $\T^f_x$ and $\T^g_x$.
With the \textit{purview}, we examine the gap in the knowledge base of $f$ that needs to be offset such that $f$ can represent $x$ more accurately.
This gap is effectively the necessary change to be made to $f$ to bring the best approximation $h(\T_1, \cdots, \T_P)$ closer to the necessary feature $\T_{P+1}$.

To assess the necessary change for the model such that $h(\T_1, \cdots, \T_P) = \T_{P+1}$, we make assumptions regarding the learned features.
Each learned feature captures a unique aspect of training data, so the features from each model are orthogonal to each other,
\begin{align}
    \T_1 \perp \T_2 \perp &\cdots \perp \T_P, \\
    \T_{P+1} \perp \T_{P+2} \perp &\cdots \perp \T_{P+Q}. \label{eq:T^g_ortho}
\end{align}
We also assume that the function $h$ for approximation is linear.
Based on the orthogonality of features in Eq.~\ref{eq:T^g_ortho} and the relationship between $\T^f_x$ and $\T^g_x$, we write
\begin{equation}
    h(\T_1, \cdots, \T_P) \perp \T_{P+2} \perp \cdots \perp \T_{P+Q} 
\end{equation}
This relationship also assures that the approximated feature is orthogonal to the $\spn$ of the rest of the features,
\begin{equation}
    \spn (\T_{P+2}, \cdots, \T_{P+Q}) \perp h(\T_1, \cdots, \T_P) \label{eq:T^g_h_ortho}
\end{equation}
To measure the necessary change, we employ gradients based on their utility in model optimization, where they correspond to the amount of change that a model requires to properly represent inputs.
Gradients have a unique property of orthogonality, which ensures that the gradients of the approximated feature would lie in the $\spn$ of the other features,
\begin{equation}
    \spn (\T_{P+2}, \cdots, \T_{P+Q}) \supset \nabla h(\T_1, \cdots, \T_P).
\end{equation}
With the linear assumption on $h$, the gradients of a linear combination of orthogonal features can be re-written as
\begin{equation}
    \spn (\T_{P+2}, \cdots, \T_{P+Q}) \supset \nabla \T_1, \cdots, \nabla \T_P.
\end{equation}
This shows that the gap in the knowledge base for the absent feature can be examined by approximating $\T_{P+2}, \cdots, \T_{P+Q}$ with the gradients of the existing features $\T^f$.
We hypothesize that the amount of necessary change for a model to represent inputs seen during inference would be more significant for inputs that differ greatly from the training data.
Equivalently, the amount of necessary change is inversely related to the \textit{effective capacity} of the model established by the training data.
Networks that are exposed to diverse training data would have enhanced \textit{effective capacity} with more generalizable features, leading to a smaller amount of required change to handle inference samples.
We propose to exploit this relationship and probe the \textit{purview} using gradients.
Similar to the use of gradients during model training, gradients can be generated during the inference of a trained model.
The obtained gradients are not applied for the actual model updates.
Because the model in question is a converged solution, the required model update would not be drastic for inputs within its knowledge base.
These gradients can be used to probe the \textit{purview} of a trained network.

%%%%%%%%%%%%%%%%%%%%%%%%%%%%%%%%%%%%%%%%%%%%%%%%%%%%%%%%%%%%%%%%
\subsection{Confounding Labels for Gradients}\label{ssec:cf_labels}

Consider an image classifier during inference.
The model will make a prediction for any given input, but the validity of model predictions remains in question due to the lack of access to labels or information on input data distribution.
This presents a challenge in utilizing gradients to probe the \textit{purview} of a trained network. 
We introduce \textit{confounding labels} to remove the dependency on ground-truth labels in gradient generation during inference.
\textit{Confounding label} $y_c$ is a label formulated by combining multiple categorical labels. 
In an image classification setting, an ordinary label consists of a single class (\ie, one-hot encodings) for a model trained to minimize cross-entropy loss, whereas a \textit{confounding label} may include multiple classes or none.
\begin{align}\label{eq:cf_label}
    y_c~\in~\big\{y : 
    &~~y~\in~ \{0,1\}^N, \nonumber\\
    &~~\textstyle\sum y_i \in \{0, \cdots, N\} \setminus \{1\} \big\}
\end{align}
Our approach focuses on the absence of relevant features in the knowledge base of a trained model, examined via gradients, to fully grasp the anomaly in the given inputs during inference. 
A \textit{confounding label} provides an unsupervised methodology to elicit a gradient response that can be utilized to probe the \textit{purview} of trained neural networks.

\begin{figure}[t]
    \centering
    \includegraphics[width=0.94\linewidth]{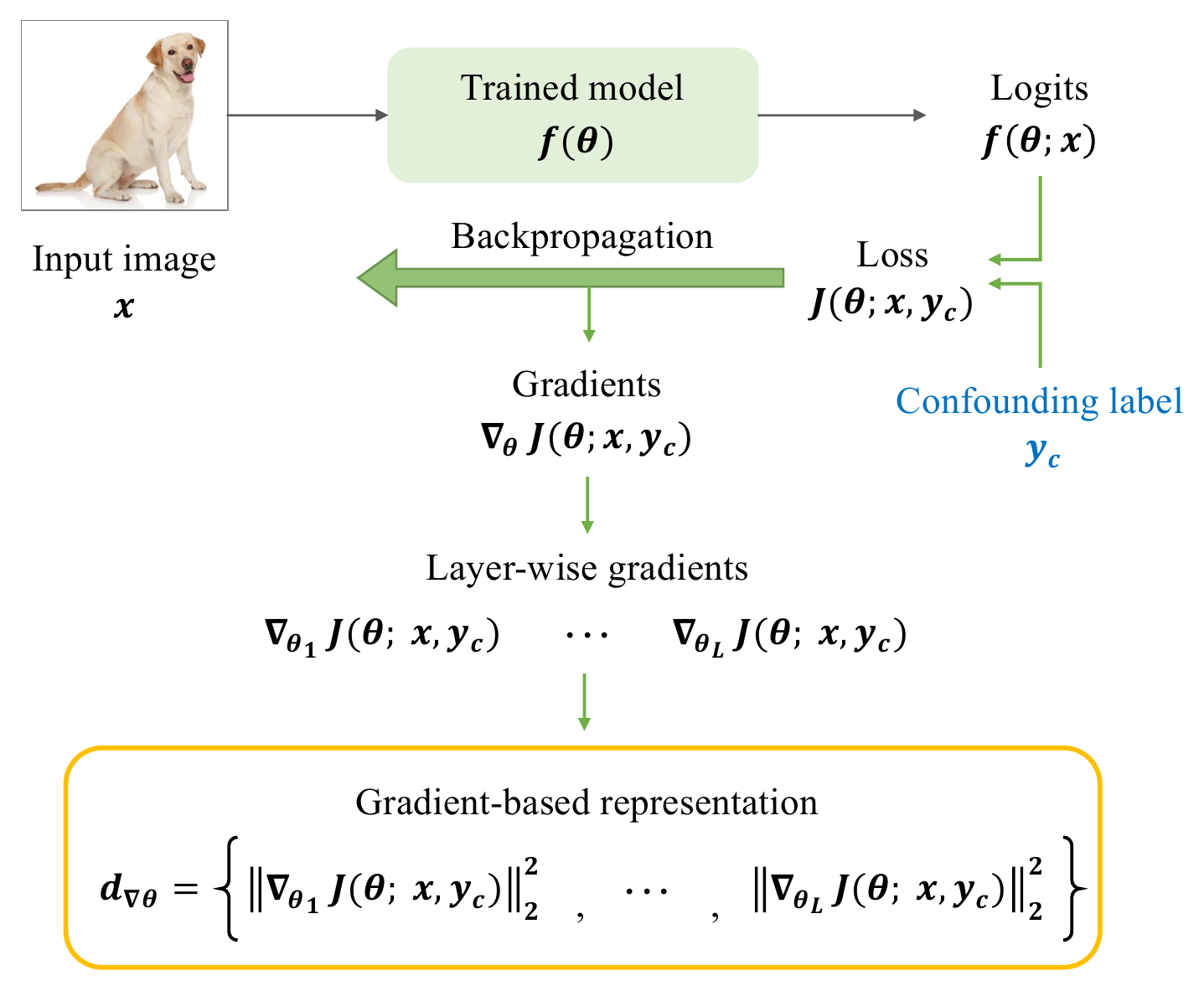}\vspace{-1mm}
    \caption{Generating gradient-based representations with \textit{confounding labels}.}\label{fig:gradient_generation}\vspace{-2mm}
\end{figure}

We discuss the framework for collecting gradient-based representations with \textit{confounding labels} during inference in Fig.~\ref{fig:gradient_generation}. 
We utilize the binary cross-entropy loss between the logits and a \textit{confounding label}, where $\hat{y_i}$ is the predicted probability for class $i$ and $y_{c,i}$ is the true probability represented by the \textit{confounding label}.
\begin{equation} \label{eq:bce} \small
    J(\theta)= -\frac{1}{N} \sum^{N-1}_{i=0} \left(y_{c,i} \cdot \log\left(\hat{y_i}\right) + \left(1 - y_{c,i}\right)\cdot\log\left(1-\hat{y_i}\right)\right)
\end{equation}
With backpropagation of the loss, gradients are generated at each set of model parameters (\ie, the weight and bias parameters of the network layers). 
While any form of gradient that preserves its magnitude would be valid, we measure the squared $L_{2}$-norm for each parameter set and concatenate them to represent the given input. 
The obtained gradient-based representation has the following form:
\begin{equation}
  \begin{gathered}
    \big[~\|\nabla J_{\theta_0}(\theta; x, y)\|^2_2~,~\cdots~,~\|\nabla J_{\theta_{L-1}}(\theta; x, y)\|^2_2~\big],
  \end{gathered}
\end{equation}
where $L$ is the number of layers or parameter sets in the given network.
We highlight that the gradient generation process involves no hyperparameters compared with other approaches that distinguish between in-distribution and OOD.

\begin{figure*}[ht]
    \centering
    \begin{subfigure}[b]{.75\textwidth}\centering
        \includegraphics[width=\textwidth]{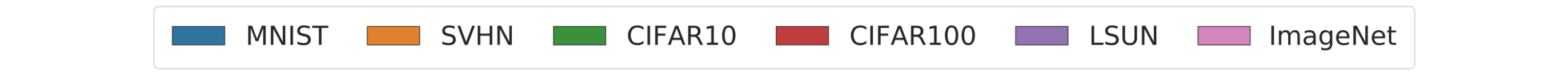}\vspace{1mm}
    \end{subfigure}
    \begin{subfigure}[c]{.41\textwidth}\centering
        \includegraphics[height=4.65cm]{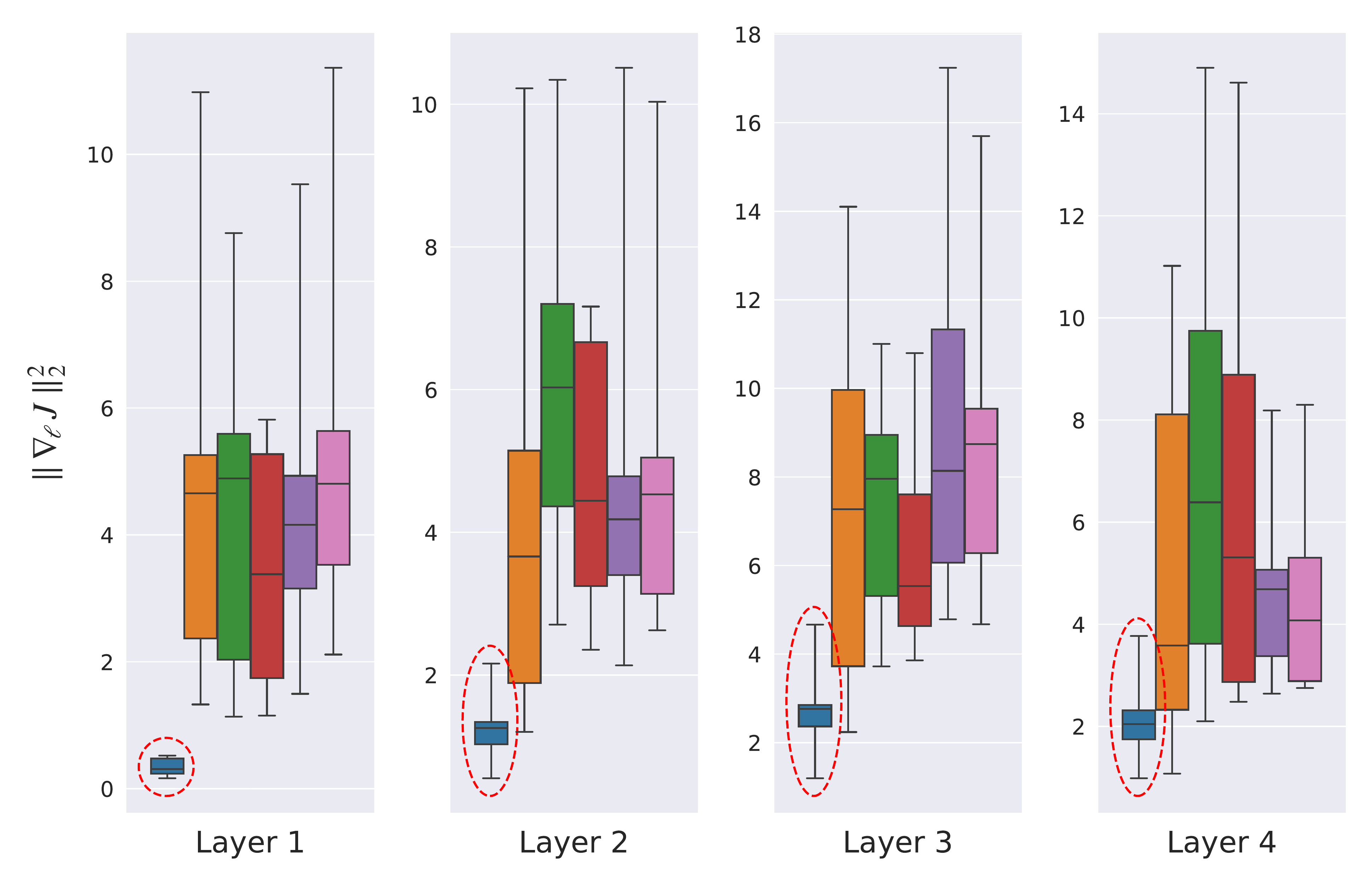}
        \subcaption{Gradients}
        \label{sfig:gradient_norm}
    \end{subfigure}
    \begin{subfigure}[c]{.45\textwidth}\centering
        \vrule \includegraphics[height=4.65cm]{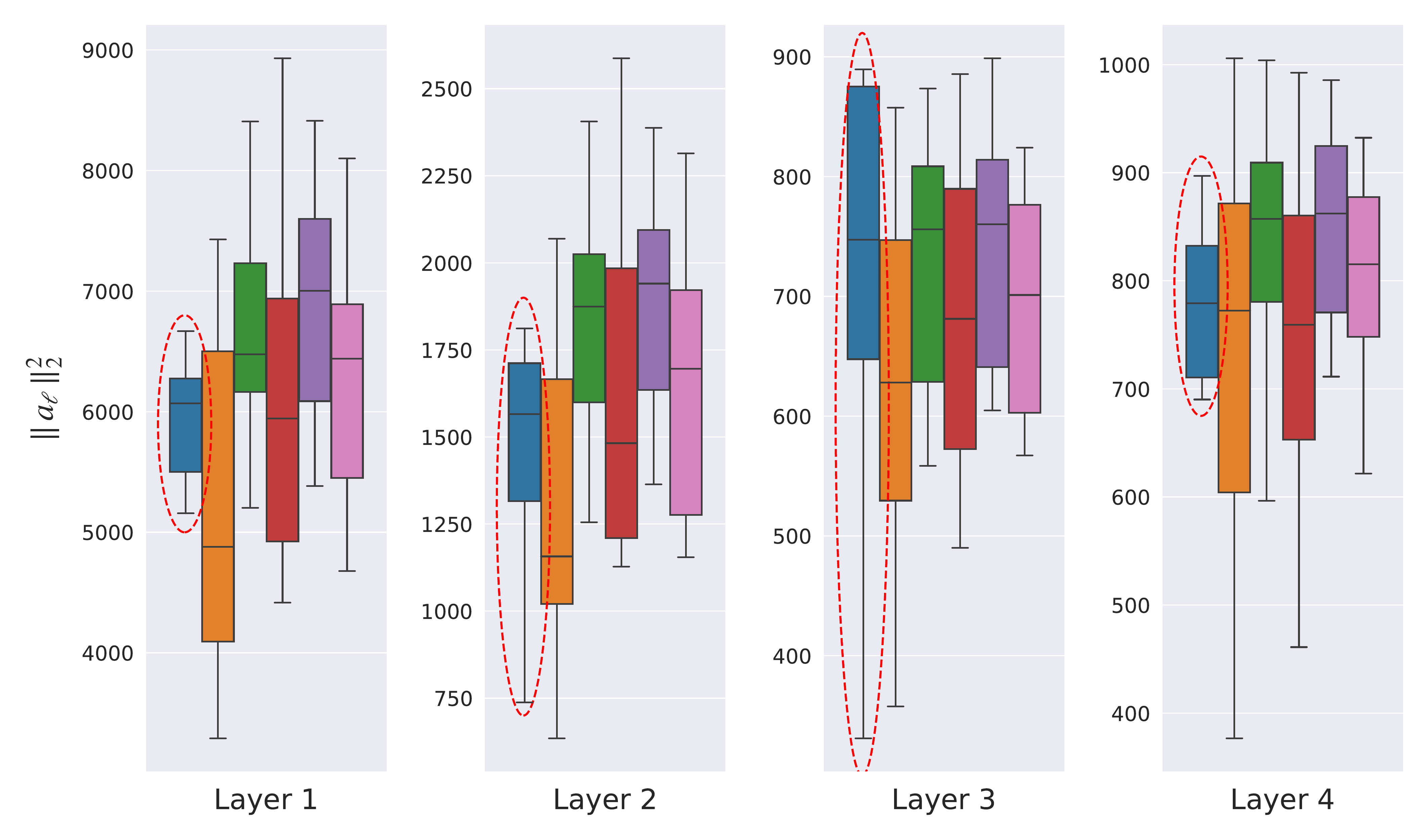}
        \subcaption{Activations}
        \label{sfig:activ_norm}
    \end{subfigure}
    \begin{subfigure}[c]{.13\textwidth}\centering
        \vrule \includegraphics[height=4.5cm]{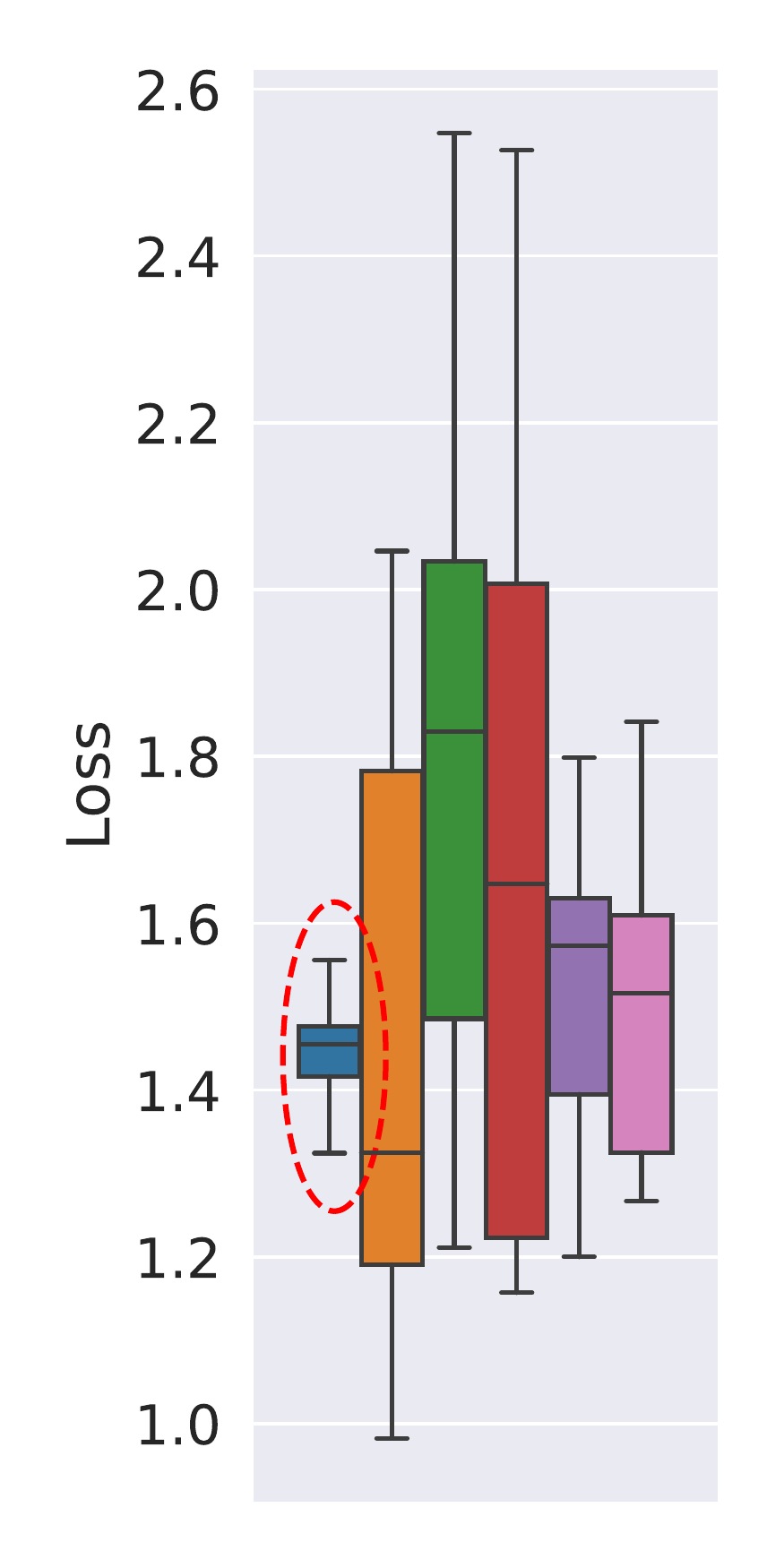}\vspace{0.5mm}
        \subcaption{Loss}
        \label{sfig:loss}
    \end{subfigure}
\vspace{1mm}
\caption{Comparison between the magnitudes of layer-wise gradients $\nabla_\ell J$, layer-wise activations $a_\ell$, and loss in distinguishing in-distribution and OOD datasets. Each plot shows distributions of per-class average values of considered attributes. The layer-wise values are obtained from a convolutional layer in each residual block of the ResNet-18 model.}
\end{figure*}\vspace{1mm}

%%%%%%%%%%%%%%%%%%%%%%%%%%%%%%%%%%%%%%%%%%%%%%%%%%%%%%%%%%%%%%%%
\subsubsection{Gradients vs. Activations vs. Loss}\label{sssec:grad_activ_loss}

We demonstrate the effectiveness of gradients obtained with \textit{confounding labels} compared to activations and loss values in differentiating anomalous inputs.
In an out-of-distribution detection setup, we show the disparity between the features learned during training and the features necessary to represent test images of both in-distribution and OOD.
The key idea is that the model will require more significant updates in its feature set to handle OOD samples than to handle in-distribution samples.
Based on our hypothesis, the gradient magnitudes obtained from the OOD samples should be larger than those of the in-distribution samples.
As discussed in Section~\ref{sec:literature}, most OOD detection approaches use activation-based measures.
We will show that gradients can capture distributional shifts more effectively than activations and loss values.

For the demonstration, a ResNet classifier~\cite{he2016resnet} is trained with MNIST~\cite{lecun1998mnist} and used to generate gradients with \textit{confounding labels} on the test sets of MNIST, SVHN~\cite{netzer2011svhn}, CIFAR-10/100~\cite{krizhevsky2009cifar}, LSUN~\cite{yu2015lsun}, and ImageNet~\cite{deng2009imagenet}, where the last two are resized subsets of their original versions provided by Liang~\etal~\cite{liang2018odin}.
We employ a network architecture that is sufficiently large for all datasets. 
We collect the squared $L_{2}$-norm of the layer-wise gradients and activations, as well as the loss values, and visualize their distributions in Fig.~\ref{sfig:gradient_norm}.
For gradient generation, we utilize a \textit{confounding label} that combines one-hot encodings of all classes (\ie, all-hot encoding). 
For visualization, we select a convolutional layer from each residual block of the ResNet architecture.
The distributions of the gradient and activation magnitudes and loss values for the in-distribution dataset are highlighted by red circles in each plot for clarity. 
The separation in the ranges of gradient magnitudes between the in-distribution and OOD datasets is more evident in some parts of the network because each layer captures information about different aspects of the given inputs. 
Nevertheless, we observe a sharp distinction based on the \textit{purview} of the model, with smaller gradient magnitudes for the in-distribution datasets and significantly larger magnitudes for the OOD datasets throughout the network layers. 
Given that the in-distribution dataset is the simplest of all considered datasets, the learned features are insufficient to characterize the OOD samples, leading to larger \textit{purview} and thus larger gradient magnitudes.
In contrast, the activations in Fig.~\ref{sfig:activ_norm} show apparent overlaps in the magnitude ranges between the in-distribution and OOD datasets throughout the network.
This supports our intuition that gradients can capture the anomaly (\ie, distributional shift in the case of OOD) in inputs better than activations based on the unfamiliar aspects of the given inputs from the model's perspective.
Similar to the activations, we observe the apparent overlap of the loss values across different datasets, as shown in Fig.~\ref{sfig:loss}. 
Loss and gradients are intertwined in the process of backpropagation, but the loss is determined based on the last layer of activation from the network and is limited to a single value per sample.
On the contrary, the gradients have the same dimensions as their corresponding parameter sets, preserving more information about the current state of the model and the necessary adjustment for better representation of the given inputs.
The overlap in the loss value range between the in-distribution and OOD datasets shows that the loss alone is inadequate for differentiating anomalous inputs from familiar inputs from the model's perspective.
Overall, gradients are more effective in characterizing anomalies in inputs than activations or loss values.

%%%%%%%%%%%%%%%%%%%%%%%%%%%%%%%%%%%%%%%%%%%%%%%%%%%%%%%%%%%%%%%%
\subsubsection{Data-Dependent Capacity and Purview}\label{sssec:data_dependent}

%%%%%%%%%%%%%%%%%%%%%%%%%%%% Purview %%%%%%%%%%%%%%%%%%%%%%%%%%%% 
\begin{figure*}[ht]
    \centering
    \begin{subfigure}[t]{\linewidth}\centering
        \includegraphics[height=4.6cm]{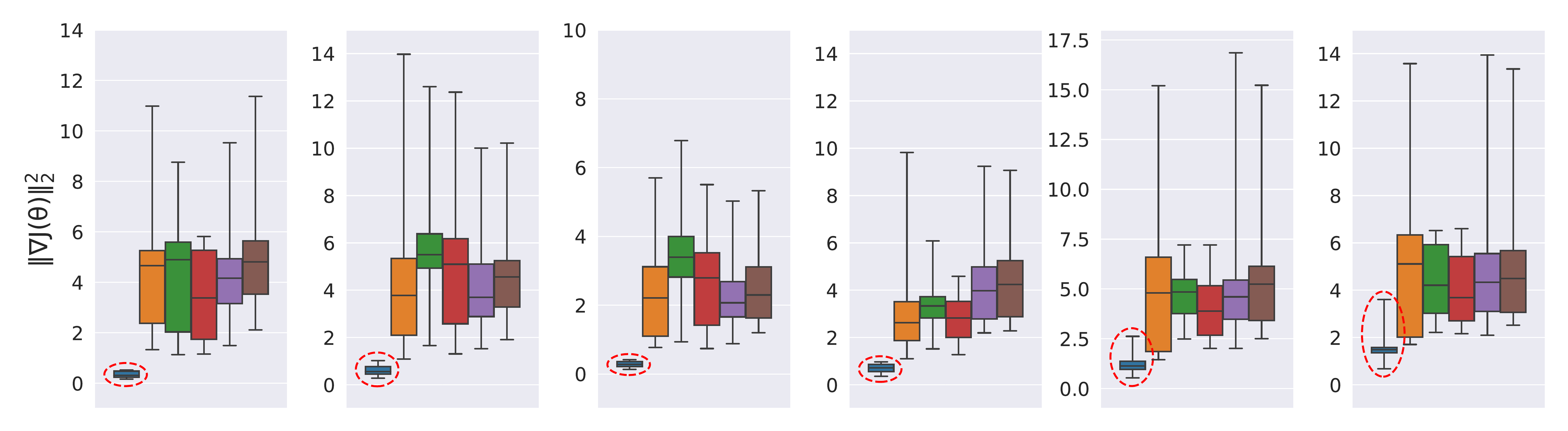}
        \subcaption{In-distribution dataset: MNIST}
        \label{sfig:grad_mnist}
    \end{subfigure}\hfill
    \begin{subfigure}[t]{\linewidth}\centering
        \includegraphics[height=4.6cm]{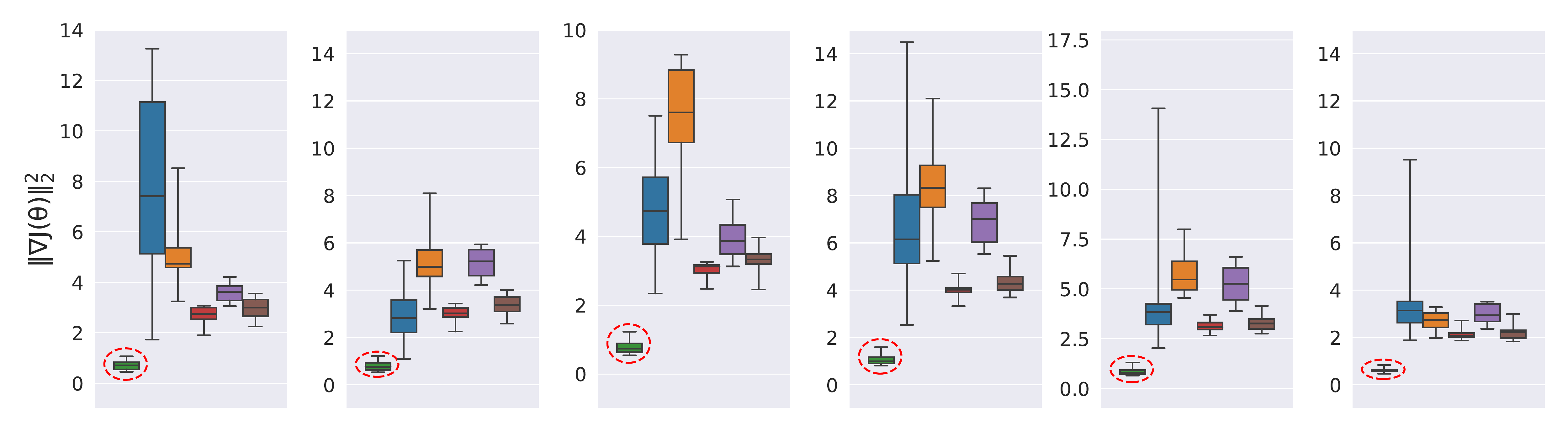}
        \subcaption{In-distribution dataset: CIFAR-10}
        \label{sfig:grad_cifar10}
    \end{subfigure}\hfill
    \begin{subfigure}[t]{\linewidth}\centering
        \includegraphics[height=4.6cm]{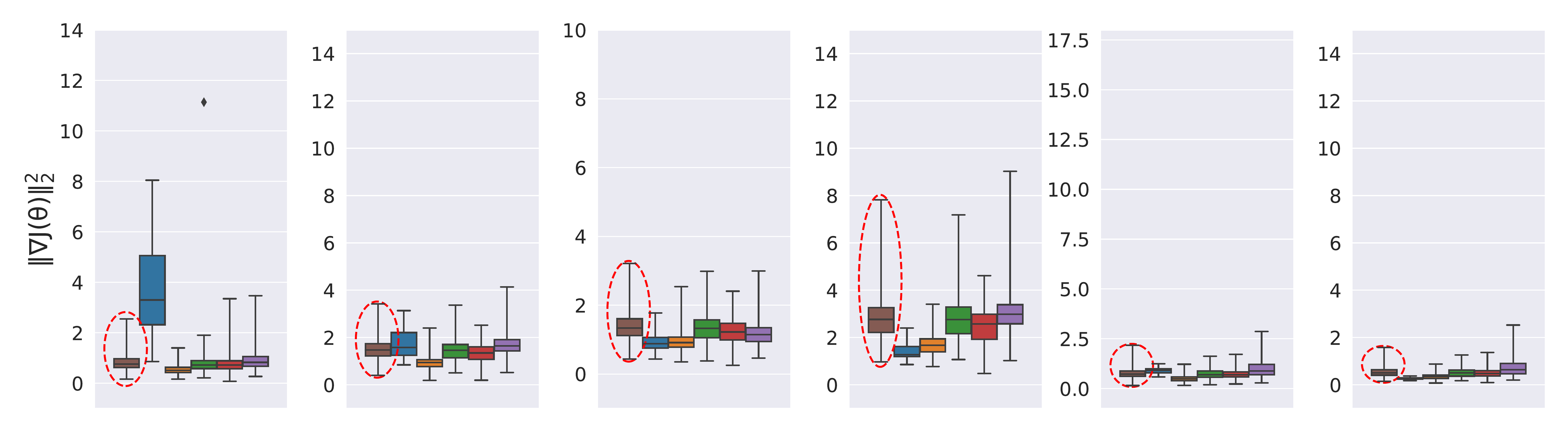}
        \subcaption{In-distribution dataset: ImageNet}
        \label{sfig:grad_tinyimagenet}
    \end{subfigure}
    \begin{subfigure}[b]{\linewidth}\centering
    \vspace{2mm}
        \includegraphics[width=0.7\linewidth]{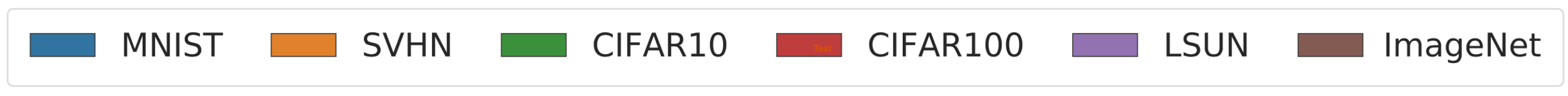}
    \end{subfigure}\hfill
\caption{Comparison of gradient magnitudes obtained from models trained with different datasets. In-distribution datasets are highlighted with red dotted circles. The model trained on the simplest dataset (\ie, MNIST) shows the largest required model updates to represent OOD samples accurately, and the distinctions in the gradient magnitudes between in-distribution datasets and OOD datasets decrease as the complexity of in-distribution dataset increases.}
\label{fig:gradient_model_scope}
\end{figure*}

We demonstrate the effectiveness of our gradient-based approach in probing the \textit{purview} of neural networks trained with data of varying levels of complexity.
In particular, we analyze the asymmetry in the data-dependent \textit{effective capacity} and the \textit{purview} observed in the gradients.
The core of this analysis is to show that a model trained on simpler data would have lower \textit{effective capacity} and more room for improvement in its feature set (\ie, \textit{purview}), leading to larger gradient responses observed throughout the network.
The experiment follows the setup of the OOD detection experiment introduced in the previous section, with the exception of the in-distribution dataset for each model.
We utilize training datasets of various complexities: MNIST, CIFAR-10, and ImageNet in ascending order of complexity in terms of the represented types and numbers of class categories.
The magnitudes of the gradients obtained with the \textit{confounding labels} are shown in Fig.~\ref{fig:gradient_model_scope} to compare the models with different degrees of \textit{effective capacity}.
Each row represents the gradient magnitudes obtained from a model trained with the specified dataset, and each column corresponds to the gradient magnitudes from a convolutional layer of the model. 
As shown in Fig.~\ref{sfig:grad_mnist}, the model trained on MNIST (\ie, the dataset with the lowest complexity) exhibits the largest gap in gradient magnitudes between the in-distribution and OOD datasets. 
It captures larger model \textit{purview} in response to the OOD samples, which is attributed to the limited data-dependent \textit{effective capacity}.
In contrast, the increase in the complexity of the in-distribution datasets leads to smaller gaps in the gradient magnitudes between the in-distribution and OOD, as seen in Fig.~\ref{sfig:grad_cifar10} and \ref{sfig:grad_tinyimagenet}. 
This indicates that by exposing models to datasets of higher complexity in training and enhancing the data-dependent \textit{effective capacity}, the learned feature sets are sufficient to capture the characteristics of the OOD samples.
However, models with less extensive feature sets would require more significant amounts of model updates to accurately represent the OOD samples.
This highlights the value of our data-dependent perspective on model capacity, which allows for learned feature-based analysis.
 
%%%%%%%%%%%%%%%%%%%%%%%%%%%%%%%%%%%%%%%%%%%%%%%%%%%%%%%%%%%%%%%%%%%%%%%%
%%%%%%%%%%%%%%%%%%%%%%%%%%%%%%%%%%%%%%%%%%%%%%%%%%%%%%%%%%%%%%%%%%%%%%%%
\section{Experiments}\label{sec:experiments}

%%%%%%%%%%%% EXP : OOD DETECTION %%%%%%%%%%%%
\begin{table*}[t]
\footnotesize % "\footnotesize" not needed here
\captionsetup{width=.98\linewidth}
\caption{Out-of-distribution detection results. All values are in percentages and the best results are highlighted in bold.}
\label{table:ood}
\begin{tabularx}{\linewidth}{@{\hspace{1mm}}l@{\hspace{2mm}}l@{\hspace{2mm}}c@{\hspace{4mm}}c@{\hspace{4mm}}c@{\hspace{1mm}}}%{@{}c*{5}{c} c @{}}
    \toprule
    \multicolumn{2}{c}{Dataset Distribution} & \multicolumn{3}{c}{Baseline~\cite{hendrycks2016baseline} / ODIN~\cite{liang2018odin} / Mahalanobis~\cite{lee2018mahalanobis} / Energy~\cite{liu2020energyOOD} / GradNorm~\cite{huang2021gradnorm} / \textit{Purview}} \\
    \cmidrule(l){3-5}
    In  &  Out & Detection Accuracy & AUROC & AUPR \\
    \midrule
SVHN            & CIFAR-10      & 86.61 / 86.25 / 88.84 / - / - / \textbf{97.55} & 91.90 / 91.76 / 95.05 / 98.85 / 97.99 / \textbf{99.74} & 92.12 / 92.38 / 90.25 / \textbf{99.57} / 96.62 / 97.76 \\
(ResNet)        & ImageNet      & 90.22 / 90.31 / 96.17 / - / - / \textbf{97.71} & 94.78 / 95.08 / 99.23 / 98.18 / 98.07 / \textbf{99.72} & 94.76 / 95.53 / 98.17 / \textbf{99.30} / 96.90 / 97.65 \\
                & LSUN          & 89.14 / 89.13 / 97.50 / - / - / \textbf{98.82} & 94.12 / 94.52 / 99.54 / 97.72 / 98.10 / \textbf{99.91} & 94.32 / 95.20 / 98.84 / 99.09 / 96.80 / \textbf{99.16} \\
                & CIFAR-100     & 79.41 / 79.58 / 89.69 / - / - / \textbf{96.49} & 81.28 / 81.32 / 95.45 / 97.61 / 97.30 / \textbf{99.39} & 80.86 / 80.84 / 85.04 / \textbf{99.06} / 95.85 / 95.70 \\
                & ImageNet(F)   & 80.62 / 80.72 / 89.55 / - / - / \textbf{96.93} & 82.44 / 82.46 / 95.79 / 98.12 / 98.01 / \textbf{99.56} & 81.59 / 81.56 / 86.64 / \textbf{99.28} / 96.88 / 96.52 \\
                & LSUN(F)       & 80.04 / 80.19 / 90.46 / - / - / \textbf{97.41} & 81.52 / 81.61 / 96.14 / 97.44 / 97.80 / \textbf{99.71} & 80.93 / 80.94 / 85.39 / \textbf{98.96} / 96.48 / 97.55 \\ \midrule
CIFAR-10        & SVHN          & 87.71 / 86.80 / 91.95 / - / - / \textbf{98.44} & 93.51 / 94.51 / 97.10 / 95.72 / 59.43 / \textbf{99.90} & 94.60 / 95.20 / 96.12 / 97.26 / 56.56 / \textbf{99.99} \\
(ResNet)        & ImageNet      & 84.93 / 91.16 / \textbf{97.45} / - / - / 90.97 & 91.50 / 96.93 / \textbf{99.68} / 84.95 / 60.69 / 96.88 & 91.02 / 96.85 / \textbf{99.60} / 84.60 / 60.18 / 97.20 \\
                & LSUN          & 85.58 / 91.91 / \textbf{98.60} / - / - / 98.59 & 92.09 / 97.51 / 99.86 / 92.39 / 70.38 / \textbf{99.90} & 91.71 / 97.51 / 99.82 / 95.40 / 69.54 / \textbf{99.91} \\
                & CIFAR-100     & 81.94 / 80.53 / \textbf{82.28} / - / - / 78.50 & 87.71 / 85.88 / \textbf{89.52} / 83.28 / 58.94 / 85.78 & 86.50 / 84.58 / \textbf{89.22} / 88.02 / 58.79 / 83.28 \\
                & ImageNet(F)   & 84.11 / 82.61 / \textbf{85.63} / - / - / 81.58 & 89.60 / 88.30 / \textbf{92.59} / 86.41 / 62.22 / 89.02 & 88.27 / 86.86 / \textbf{92.44} / 90.47 / 61.57 / 86.76 \\
                & LSUN(F)       & 84.47 / 83.37 / \textbf{84.75} / - / - / 84.28 & 89.92 / 89.00 / 91.70 / 87.94 / 64.12 / \textbf{92.23} & 88.79 / 87.78 / \textbf{91.18} / 91.11 / 63.88 / 91.00 \\ \midrule
SVHN            & CIFAR-10      & 86.61 / 86.25 / 96.50 / - / - / \textbf{98.06} & 91.90 / 91.76 / 99.02 / 90.63 / 37.76 / \textbf{99.79} & 92.12 / 92.38 / 94.40 / 94.30 / 46.65 / \textbf{98.16} \\
(DenseNet)      & ImageNet      & 90.22 / 90.31 / \textbf{98.89} / - / - / 98.01 & 94.78 / 95.08 / \textbf{99.84} / 87.51 / 40.28 / 99.77 & 94.76 / 95.53 / \textbf{99.37} / 92.15 / 48.76 / 98.18 \\
                & LSUN          & 89.14 / 89.13 / \textbf{99.49} / - / - / 99.46 & 94.12 / 94.52 / 99.92 / 83.45 / 38.85 / \textbf{99.95} & 94.32 / 95.20 / 99.54 / 89.33 / 47.05 / \textbf{99.61} \\
                & CIFAR-100     & 86.59 / 85.85 / 96.49 / - / - / \textbf{97.12} & 91.42 / 90.95 / 99.14 / 86.09 / 37.66 / \textbf{99.63} & 91.60 / 91.73 / 95.52 / 90.94 / 47.09 / \textbf{97.09} \\
                & ImageNet(F)   & 87.28 / 86.50 / 96.26 / - / - / \textbf{97.35} & 92.18 / 91.68 / 99.15 / 87.06 / 39.04 / \textbf{99.69} & 92.33 / 92.42 / 95.77 / 91.58 / 48.08 / \textbf{97.64} \\
                & LSUN(F)       & 86.63 / 86.40 / 96.07 / - / - / \textbf{98.57} & 91.69 / 91.69 / 99.04 / 86.01 / 42.45 / \textbf{99.83} & 91.93 / 92.34 / 94.63 / 91.34 / 49.79 / \textbf{98.79} \\ \midrule
CIFAR-10        & SVHN          & 87.71 / 86.80 / 95.75 / - / - / \textbf{98.89} & 93.51 / 94.51 / 98.96 / 89.33 / 88.75 / \textbf{99.95} & 94.60 / 95.20 / 97.21 / 94.01 / 83.94 / \textbf{99.99} \\
(DenseNet)      & ImageNet      & 84.93 / 91.16 / \textbf{96.83} / - / - / 89.47 & 91.50 / 96.93 / \textbf{99.45} / 86.17 / 74.25 / 95.99 & 91.02 / 96.85 / \textbf{99.29} / 87.20 / 72.69 / 95.92 \\
                & LSUN          & 85.58 / 91.91 / 98.08 / - / - / \textbf{99.24} & 92.09 / 97.51 / 99.74 / 92.77 / 84.15 / \textbf{99.96} & 91.71 / 97.51 / 99.72 / 95.57 / 83.71 / \textbf{99.96} \\
                & CIFAR-100     & 81.12 / \textbf{81.32} / 65.75 / - / - / 74.05 & 87.92 / \textbf{88.89} / 71.90 / 83.32 / 71.91 / 82.12 & 87.08 / 88.36 / 71.61 / \textbf{89.17} / 70.39 / 80.57 \\
                & ImageNet(F)   & 82.22 / \textbf{83.35} / 69.68 / - / - / 81.84 & 88.89 / \textbf{90.51} / 77.35 / 87.06 / 72.72 / 89.95 & 88.16 / 90.02 / 77.00 / \textbf{92.00} / 71.01 / 88.65 \\
                & LSUN(F)       & 81.75 / 81.89 / 75.03 / - / - / \textbf{90.28} & 88.42 / 89.57 / 82.05 / 86.77 / 68.12 / \textbf{96.39} & 87.64 / 89.19 / 81.21 / 92.01 / 66.49 / \textbf{96.14} \\
\bottomrule
\end{tabularx}\vspace{-2mm}
\end{table*}

In this section, we apply our gradient-based approach to probe the \textit{purview} of neural networks to detect anomalous inputs: OOD detection, adversarial detection, and corrupted input detection. 
We distinguish the anomaly based on the application and training datasets.
For OOD detection, the in-distribution datasets are referred to as normal, and the OOD datasets are anomalous.
For adversarial and corrupted input detection, clean images are considered normal, and adversarial attack images or corrupted images are considered anomalous.
From a trained classifier, we obtain gradient-based representations from the test sets of both normal and anomalous datasets using a \textit{confounding label} where all classes are positive (\ie, all-hot encoding).
For each experiment, the collected gradient representations are used to train a simple binary detector with two fully-connected layers as an anomalous input detector.
Each reported detection result is an average of 5-fold cross-validation results repeated with two different random seeds, totaling ten rounds with randomly initialized detectors for all detection experiments.
Further details regarding the implementation are provided in Section~\ref{ssec:implementation}.
In addition, we provide an ablation study on detector network architectures and training, as well as the designs of \textit{confounding labels} in Section~\ref{ssec:ablation}.
We note that our approach is not specifically devised to perform well for these applications. 
Rather, it is a byproduct of the focus on understanding the \textit{purview} of models and the apparent gap in the data-dependent capacity observed in the distribution of the gradient responses and has proven useful in such applications.

%%%%%%%%%%%%%%%%%%%%%%%%%%%%%%%%%%%%%%%%%%%%%%%%%%%%%%%%%%%%%%%%%%%%%%%%
\subsection{Out-of-Distribution Detection}\label{ssec:ood}

We apply our gradient-based approach on OOD detection using various image classification datasets.
We utilize CIFAR-10 and SVHN datasets as in-distribution.
For OOD, we consider the following additional datasets: CIFAR-100~\cite{krizhevsky2009cifar}, resized LSUN and ImageNet~\cite{liang2018odin}, and fixed LSUN and ImageNet~\cite{tack2020csi}, denoted as LSUN (FIX) and ImageNet (FIX), respectively.
ResNet-18 and DenseNet~\cite{huang2017densenet} architectures are utilized as classifiers from which gradient-based representations are obtained.
We measure the detection performance with the following evaluation metrics:
\begin{itemize}[leftmargin=4mm,itemsep=2mm,topsep=2mm]
\item \textbf{Detection accuracy} measures the maximum classification accuracy over all possible confidence thresholds $\delta$,
\begin{equation*}
    \small \operatorname*{max}_{\delta} \left\{0.5 P_{in}(q(x) \leq \delta) + 0.5 P_{out}(q(x) > \delta)  \right\},
\end{equation*}
where $q(x)$ denotes the confidence score from the detector. 
For our method, we fix the threshold $\delta$ to $0.5$.
\item \textbf{AUROC} measures the area under the receiver operating characteristic curve, which plots the true positive rate against the false positive rate obtained with varying threshold settings.
\item \textbf{AUPR} measures the area under the precision-recall curve, which plots the precision (the ratio between true positives and all positives) against the true positive rate with varying threshold values.
\end{itemize}

\begin{figure}[t]
    \centering
    \begin{subfigure}[t]{.23\linewidth}\centering
        \includegraphics[width=\linewidth]{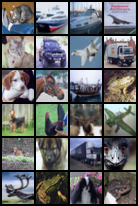}
        \subcaption{CIFAR-10}\label{sfig:dataset_cifar10}
    \end{subfigure}
    \begin{subfigure}[t]{.23\linewidth}\centering
        \includegraphics[width=\linewidth]{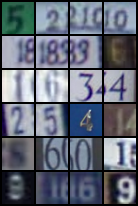}
        \subcaption{SVHN}\label{sfig:dataset_svhn}
    \end{subfigure}
    \begin{subfigure}[t]{.23\linewidth}\centering
        \includegraphics[width=\linewidth]{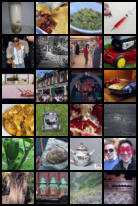}
        \subcaption{ImageNet}\label{sfig:dataset_tinyimagenet}
    \end{subfigure}
    \begin{subfigure}[t]{.23\linewidth}\centering
        \includegraphics[width=\linewidth]{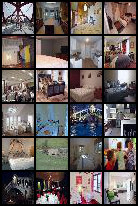}
        \subcaption{LSUN}\label{sfig:dataset_lsun}
    \end{subfigure}
\caption{Sample dataset images.}\label{fig:dataset_samples}
\vspace{-5mm}
\end{figure}

\begin{table*}[ht]
\caption{Adversarial detection results for CIFAR-10 in AUROC. For detecting unknown attacks, attacks denoted by \textit{``seen''} are used for detector training and evaluated on unknown attacks.}
\label{table:adversarial}
\centering
\begin{small}
\setlength{\tabcolsep}{6.8pt}
\begin{tabularx}{\linewidth}{cl|cccccc|cc|cc}
\toprule
\multirow{2}{*}{Model} & \multirow{2}{*}{Attacks} & \multirow{2}{*}{Baseline} & \multirow{2}{*}{LID} & \multirow{2}{*}{M(V)} & \multirow{2}{*}{M} & \multirow{2}{*}{Hu~\etal} & \multirow{2}{*}{\textit{Purview}} & \multicolumn{2}{c|}{(Seen: BIM)} & \multicolumn{2}{c}{(Seen: PGD)} \\
 &  &  &  &  &  &  &  & M & \textit{Purview} & M & \textit{Purview} \\
\midrule
\multirow{7}{*}{ResNet} 
 & FGSM & 68.89 & 90.06 & 81.69 & \textbf{99.95} & 84.06 & 93.45 & \textbf{99.88} & 90.11 & 83.97 & \textbf{91.33}\\
 & BIM & 54.09 & 99.21 & 87.09 & \textbf{100.0} & 83.65 & 96.19  & (\textbf{100.0}) & (96.19) & 74.44 & \textbf{95.89}\\
 & C\&W & 87.58 & 76.47 & 74.51  & 92.79 & 84.02 & \textbf{97.07} & 86.65 & \textbf{94.98} & 77.54 & \textbf{95.37}\\
 & PGD & 57.22 & 67.48 & 56.27  & 75.98 & 83.93 & \textbf{95.82} & 60.98 & \textbf{94.87} & (75.98) & (\textbf{95.82}) \\
 & IterLL & 82.00 & 85.17 & 62.32 & 92.10 & 84.22 & \textbf{98.17} & 81.48 & \textbf{95.10} & 92.24 & \textbf{96.42} \\
 & Semantic & 78.88 & 86.25 & 64.18 & 84.38 & 80.95 & \textbf{90.15} & \textbf{70.62} & 60.50 & \textbf{83.22} & 60.81\\
 & AutoAttack & 5.13 & 96.37 & 76.15 & 95.98 & 69.12 & \textbf{99.30} & 85.71 & \textbf{97.71} & 95.55 & \textbf{96.33} \\
 \midrule
\multirow{7}{*}{DenseNet}
 & FGSM & 75.25 & 98.23 & 86.88 & \textbf{99.97} &91.67 & 96.83  & \textbf{99.87} & 95.15 & \textbf{99.70} & 95.60 \\
 & BIM & 47.01 & \textbf{100.0} & 89.19 & \textbf{100.0} & 90.32 & 96.85  & (\textbf{100.0}) & (96.85) & \textbf{100.0} & 96.99\\
 & C\&W & 88.59 & 80.58 & 75.77 & 90.76 &93.88 & \textbf{97.05}  & 78.06 & \textbf{95.66} & 84.36 & \textbf{96.11}\\
 & PGD & 50.30 & 83.01 & 70.39  & 83.61 &93.55 & \textbf{96.77} & 75.64 & \textbf{96.71} & (83.61) & (\textbf{96.77})\\
 & IterLL & 63.68 & 83.16 & 70.17 & 77.84 &91.67 & \textbf{98.53}  & 75.68 & \textbf{95.54} & 79.39 & \textbf{96.23}\\
 & Semantic & 79.22 & 81.41 & 62.16 & 67.29 &\textbf{94.12} & 89.55  & \textbf{66.59} & 57.63 & 45.85 & \textbf{58.38}\\
 & AutoAttack &  31.93 & 95.06 & 90.15 & 97.31 &36.00 & \textbf{98.16} & \textbf{97.57} & 96.91 & 94.30 & \textbf{95.45}\\
\bottomrule
\end{tabularx}
\end{small}
\end{table*}

\noindent The results of OOD detection are reported in Table~\ref{table:ood} along with other state-of-the-art methods, including Baseline~\cite{hendrycks2016baseline}, ODIN~\cite{liang2018odin}, Mahalanobis~\cite{lee2018mahalanobis}, Energy~\cite{liu2020energyOOD}, and GradNorm~\cite{huang2021gradnorm}. 
For the Energy and GradNorm methods, the detection accuracy is omitted because they do not specifically determine the threshold values for their OOD scores for detection.
We observe that our method, shown as \textit{Purview}, outperforms all state-of-the-art methods in AUROC when SVHN is used as in-distribution in 11 out of 12 cases.
With CIFAR-10 as in-distribution, our method outperforms all others in AUROC when SVHN, LSUN, and LSUN (FIX) are used as OOD. 
Our method is particularly effective when there is a considerable difference in the complexity of the in-distribution and OOD datasets. 
We provide sample images from each dataset for the visual analysis in Fig.~\ref{fig:dataset_samples}.
ImageNet shows the most visual similarity with CIFAR-10 in terms of the types and scales of objects in the images among the three OOD datasets shown in Fig.~\ref{fig:dataset_samples}~\subref{sfig:dataset_svhn}-\subref{sfig:dataset_lsun}.
SVHN is a street-view house number dataset that is vastly different from CIFAR-10 dataset of natural images. 
LSUN is a scene-understanding dataset with scene categories for classification.
The results support our intuition behind the \textit{purview} of a trained network based on its data-dependent \textit{effective capacity}. 
The models trained on CIFAR-10 have higher \textit{effective capacity} than those trained on the simpler SVHN.
Their feature sets include more diverse features that can be utilized to approximate the unfamiliar characteristics of the OOD datasets.
On the other hand, models trained on SVHN have relatively simple features.
They require more extensive updates to handle OOD datasets, exhibiting a more apparent gap in the gradient responses and leading to a better OOD detection performance.

Similar reasoning can be applied to the inferior detection performance observed with ImageNet and CIFAR-100 as OOD.
With SVHN as in-distribution, our approach achieves the best performance in all cases except one, and a very close second in one case.
However, with CIFAR-10 as in-distribution, our approach is outperformed by up to 3.5\% AUROC for ImageNet, 3.7\% for ImageNet(FIX), and 6.8\% for CIFAR-100.
Our experiment results show that the more similar the in-distribution and OOD datasets are, the smaller the gap in the spreads of gradient magnitudes, which makes it more challenging to differentiate OOD samples from in-distribution samples. 
Overall, our gradient-based method outperforms all activation-based methods in 17 out of 24 cases in terms of AUROC.

%%%%%%%%%%%%%%%%%%%%%%%%%%%%%%%%%%%%%%%%%%%%%%%%%%%%%%%%%%%%%%%%%%%%%%%%
\subsection{Adversarial Detection}\label{ssec:adversarial}

We utilize ResNet and DenseNet classifiers trained on CIFAR-10 training set for adversarial detection. 
The test set of CIFAR-10 is utilized to generate the following adversarial attacks: fast gradient sign method (FGSM)~\cite{goodfellow2014adversarial}, basic iterative method (BIM)~\cite{kurakin2017bim_iterll}, Carlini \& Wagner attack (C\&W)~\cite{carlini2017cw}, projected gradient descent (PGD)~\cite{madry2017pgd}, iterative least-likely class method (IterLL)~\cite{kurakin2017bim_iterll}, semantic attack~\cite{hosseini2017semantic}, and AutoAttack~\cite{croce2020autoattack}. 
We utilize the pristine CIFAR-10 test set as normal samples and adversarial images as anomalous samples. 
For comparison, we employ the Baseline method, local intrinsic dimensionality (LID) scores~\cite{ma2018lid}, the Mahalanobis method, and the approach introduced by Hu~\etal~\cite{hu2019new}. 
The performance is measured in AUROC and reported in Table~\ref{table:adversarial}. 
For the Mahalanobis method (M), we report vanilla results (V), \ie, without the input pre-processing or feature ensemble, and results with both to demonstrate its dependency on the delicate calibration of hyperparameters, neither of which our approach requires.

The Mahalanobis method performs best for the adversarial attack types of FGSM and BIM. 
In contrast, our approach outperforms all state-of-the-art methods for C\&W, PGD, IterLL, Semantic and AutoAttack attacks. 
Note that the Mahalanobis method requires fine-tuning of hyperparameters for each attack type, and we report their results with the best-performing parameters selected within the range of values explored in their study. 
Interestingly, the Mahalanobis approach shows a significant drop in performance for the types of attacks that were not included in their work. 
This indicates the need for additional hyperparameter tuning outside of the suggested parameter values. 
It highlights the benefit of our approach in its simplicity of obtaining gradient-based representations without the need for pre- or post-processing. 
Our method can effectively characterize adversarial attack samples, even for attack types widely considered highly challenging to detect: C\&W, PGD, and AutoAttack.

%%%%%%%%%%%%%%%%%%%%%%%%%%%%%%%%%%%%%%%%%
%%%%%%%%%%%% DTACC: CURE-TSR %%%%%%%%%%%%
%%%%%%%%%%%%%%%%%%%%%%%%%%%%%%%%%%%%%%%%%
%%%% Corrected k-fold Paired T-Test

\begin{table*}[ht]
\scriptsize	
\caption{Corrupted input detection results for CURE-TSR dataset in detection accuracy. All values are in percentages and the best results are highlighted in bold. The results shown in colors are statistically significant, where green shows \textit{purview} outperforms Mahalanobis, and red vice versa. (\textit{Lev.}: corruption severity levels, \textit{Avg.}: average scores across all levels for each corruption type.)}
\label{table:corrupted_curetsr}
\centering
\setlength{\tabcolsep}{0.9pt}
\begin{tabularx}{\linewidth}{@{\hspace{0.6ex}}c@{\hspace{0.6ex}}c@{\hspace{1.2ex}}c@{\hspace{1.2ex}}c@{\hspace{1.2ex}}c@{\hspace{1.2ex}}c@{\hspace{1.2ex}}c@{\hspace{1.2ex}}c@{\hspace{1.2ex}}c@{\hspace{1.2ex}}c@{\hspace{1.2ex}}c@{\hspace{1.2ex}}c@{\hspace{1.2ex}}c@{\hspace{0.6ex}}}
\toprule
 & \multicolumn{12}{c@{}}{Mahalanobis~\cite{lee2018mahalanobis} / \textit{Purview}} \\
\cmidrule(l){2-13}
Lev.  & Decolor & LensBlur & CodecError & Darkening & DirtyLens & Exposure & GaussianBlur   & Noise    & Rain       & Shadow    & Snow      & Haze \\
\midrule
1 & \textbf{62.82} / 59.30 & 58.77 / \textbf{66.50} & \cellcolor[HTML]{B7E1CD}56.22 / \textbf{72.14} & \cellcolor[HTML]{B7E1CD}67.84 / \textbf{79.75} & \textbf{53.37} / 50.45 & \cellcolor[HTML]{B7E1CD}70.54 / \textbf{80.03} & \cellcolor[HTML]{B7E1CD}61.69 / \textbf{75.16} & \textbf{53.67} / 50.22 & \cellcolor[HTML]{B7E1CD}74.81 / \textbf{93.83} & \cellcolor[HTML]{F4C7C3}\textbf{62.37} / 56.48 & 51.35 / \textbf{56.15} & \cellcolor[HTML]{B7E1CD}70.39 / \textbf{89.55} \\
2 & 70.01 / \textbf{73.56} & \cellcolor[HTML]{B7E1CD}67.54 / \textbf{79.21} & \cellcolor[HTML]{B7E1CD}59.00 / \textbf{74.63} & \cellcolor[HTML]{B7E1CD}74.21 / \textbf{96.51} & 51.27 / \textbf{57.07} & \cellcolor[HTML]{F4C7C3}\textbf{99.40} / 91.24 & \cellcolor[HTML]{B7E1CD}70.99 / \textbf{78.31} & \textbf{62.07} / 59.51 & \cellcolor[HTML]{B7E1CD}89.81 / \textbf{96.24} & 57.72 / \textbf{62.41} & \cellcolor[HTML]{B7E1CD}64.47 / \textbf{72.26} & 95.50 / \textbf{96.48} \\
3 & \cellcolor[HTML]{B7E1CD}68.37 / \textbf{81.92} & 81.78 / \textbf{84.69} & \cellcolor[HTML]{B7E1CD}55.62 / \textbf{79.30} & \cellcolor[HTML]{B7E1CD}98.20 / \textbf{99.45} & \cellcolor[HTML]{B7E1CD}57.80 / \textbf{74.99} & \cellcolor[HTML]{F4C7C3}\textbf{99.33} / 95.70 & \textbf{87.48} / 86.63 & \cellcolor[HTML]{B7E1CD}59.37 / \textbf{73.94} & \cellcolor[HTML]{B7E1CD}93.10 / \textbf{96.30} & \cellcolor[HTML]{B7E1CD}54.27 / \textbf{65.64} & 82.01 / \textbf{82.71} & \cellcolor[HTML]{F4C7C3}\textbf{100.0} / 99.18 \\
4 & \cellcolor[HTML]{B7E1CD}78.71 / \textbf{87.78} & \textbf{86.43} / 86.72 & \cellcolor[HTML]{B7E1CD}55.55 / \textbf{82.96} & \textbf{100.0} / 99.93 & \cellcolor[HTML]{B7E1CD}67.54 / \textbf{83.49} & \textbf{99.70} / 97.60 & \cellcolor[HTML]{B7E1CD}80.28 / \textbf{87.49} & \cellcolor[HTML]{B7E1CD}63.49 / \textbf{83.31} & \cellcolor[HTML]{B7E1CD}94.53 / \textbf{96.35} & \cellcolor[HTML]{B7E1CD}51.35 / \textbf{69.00} & \cellcolor[HTML]{F4C7C3}\textbf{95.50} / 88.89 & \textbf{100.0} / 98.41 \\
5 & \cellcolor[HTML]{B7E1CD}80.88 / \textbf{90.34} & \textbf{90.33} / 88.68 & \cellcolor[HTML]{B7E1CD}56.07 / \textbf{84.27} & \textbf{100.0} / 99.79 & \textbf{95.28} / 94.48 & \textbf{100.0} / 99.51 & \cellcolor[HTML]{F4C7C3}\textbf{92.28} / 89.63 & \cellcolor[HTML]{B7E1CD}75.64 / \textbf{87.68} & 96.63 / \textbf{97.75} & \cellcolor[HTML]{B7E1CD}63.34 / \textbf{72.11} & \cellcolor[HTML]{F4C7C3}\textbf{99.33} / 91.77 & \textbf{100.0} / 98.79 \\ 
\midrule
Avg. & \cellcolor[HTML]{B7E1CD}72.16 / \textbf{78.58} & \cellcolor[HTML]{B7E1CD}76.97 / \textbf{81.16} & \cellcolor[HTML]{B7E1CD}56.49 / \textbf{78.66} & \cellcolor[HTML]{B7E1CD}88.05 / \textbf{95.08} & \cellcolor[HTML]{B7E1CD}65.05 / \textbf{72.09} & \textbf{93.79} / 92.82 & \cellcolor[HTML]{B7E1CD}78.55 / \textbf{83.44} & \cellcolor[HTML]{B7E1CD}62.85 / \textbf{70.93} & \cellcolor[HTML]{B7E1CD}89.78 / \textbf{96.10} & \cellcolor[HTML]{B7E1CD}57.81 / \textbf{65.13} & 78.53 / \textbf{78.36} & \cellcolor[HTML]{B7E1CD}93.18 / \textbf{96.48} \\
\bottomrule
\end{tabularx}
\end{table*}

%%%%%%%%%%%%%%%%%%%%%%%%%%%%%%%%%%%%%%%%%%%
%%%%%%%%%%%% DTACC: CIFAR-10-C %%%%%%%%%%%% 
%%%%%%%%%%%%%%%%%%%%%%%%%%%%%%%%%%%%%%%%%%%

\begin{table*}[ht]
\scriptsize
\caption{Corrupted input detection results for CIFAR-10-C dataset in detection accuracy. All values are in percentages and the best results are highlighted in bold. The results shown in colors are statistically significant, where green shows \textit{purview} outperforms Mahalanobis, and red vice versa.}
\label{table:corrupted_cifar10c}
\centering
\setlength{\tabcolsep}{3pt}
\begin{tabularx}{\linewidth}{cc@{\hspace{4.9ex}}c@{\hspace{4.9ex}}c@{\hspace{4.9ex}}c@{\hspace{4.9ex}}c@{\hspace{4.9ex}}c@{\hspace{4.9ex}}c@{\hspace{4.9ex}}c@{\hspace{4.9ex}}c}
\toprule
 & \multicolumn{9}{c@{}}{Mahalanobis~\cite{lee2018mahalanobis} / \textit{Purview}} \\
\cmidrule(l){2-10}
Levels  & GaussianNoise & ShotNoise & ImpulseNoise & SpeckleNoise & GaussianBlur & DefocusBlur & GlassBlur & MotionBlur & ZoomBlur  \\
\midrule
1 & \cellcolor[HTML]{B7E1CD}90.88 / \textbf{99.17} & \cellcolor[HTML]{B7E1CD}88.93 / \textbf{98.86} & \cellcolor[HTML]{B7E1CD}90.65 / \textbf{99.07} & \cellcolor[HTML]{B7E1CD}88.75 / \textbf{98.95} & \cellcolor[HTML]{B7E1CD}87.78 / \textbf{99.01} & \cellcolor[HTML]{B7E1CD}87.90 / \textbf{99.16} & \cellcolor[HTML]{B7E1CD}96.95 / \textbf{99.61} & \cellcolor[HTML]{B7E1CD}94.30 / \textbf{99.67} & \cellcolor[HTML]{B7E1CD}96.18 / \textbf{99.82}  \\
2 & \cellcolor[HTML]{B7E1CD}94.93 / \textbf{99.44} & \cellcolor[HTML]{B7E1CD}91.20 / \textbf{99.11} & \cellcolor[HTML]{B7E1CD}93.50 / \textbf{99.55} & \cellcolor[HTML]{B7E1CD}91.83 / \textbf{99.18} & \cellcolor[HTML]{B7E1CD}96.33 / \textbf{99.78} & \cellcolor[HTML]{B7E1CD}92.33 / \textbf{99.57} & \cellcolor[HTML]{B7E1CD}96.70 / \textbf{99.61} & \cellcolor[HTML]{B7E1CD}97.15 / \textbf{99.85} & \cellcolor[HTML]{B7E1CD}97.13 / \textbf{99.86} \\
3 & \cellcolor[HTML]{B7E1CD}97.10 / \textbf{99.66} & \cellcolor[HTML]{B7E1CD}94.95 / \textbf{99.37} & \cellcolor[HTML]{B7E1CD}95.73 / \textbf{99.64} & \cellcolor[HTML]{B7E1CD}93.30 / \textbf{99.37} & \cellcolor[HTML]{B7E1CD}98.30 / \textbf{99.93} & \cellcolor[HTML]{B7E1CD}96.30 / \textbf{99.81} & \cellcolor[HTML]{B7E1CD}96.05 / \textbf{99.80} & \cellcolor[HTML]{B7E1CD}98.10 / \textbf{99.94} & \cellcolor[HTML]{B7E1CD}98.05 / \textbf{99.96}  \\
4 & \cellcolor[HTML]{B7E1CD}97.75 / \textbf{99.65} & \cellcolor[HTML]{B7E1CD}96.45 / \textbf{99.55} & \cellcolor[HTML]{B7E1CD}98.18 / \textbf{99.77} & \cellcolor[HTML]{B7E1CD}94.93 / \textbf{99.51} & \cellcolor[HTML]{B7E1CD}99.25 / \textbf{99.97} & \cellcolor[HTML]{B7E1CD}98.38 / \textbf{99.90} & \cellcolor[HTML]{B7E1CD}97.40 / \textbf{99.76} & \cellcolor[HTML]{B7E1CD}98.05 / \textbf{99.92} & \cellcolor[HTML]{B7E1CD}98.40 / \textbf{99.93} \\
5 & \cellcolor[HTML]{B7E1CD}98.15 / \textbf{99.67} & \cellcolor[HTML]{B7E1CD}97.53 / \textbf{99.53} & \cellcolor[HTML]{B7E1CD}99.08 / \textbf{99.85} & \cellcolor[HTML]{B7E1CD}96.05 / \textbf{99.52} & \cellcolor[HTML]{B7E1CD}99.18 / \textbf{99.95} & \cellcolor[HTML]{B7E1CD}99.53 / \textbf{99.93} & \cellcolor[HTML]{B7E1CD}96.63 / \textbf{99.73} & \cellcolor[HTML]{B7E1CD}98.60 / \textbf{99.92} & \cellcolor[HTML]{B7E1CD}98.85 / \textbf{99.92}  \\
\midrule
Average & \cellcolor[HTML]{B7E1CD}95.76 / \textbf{99.52} & \cellcolor[HTML]{B7E1CD}93.81 / \textbf{99.28} & \cellcolor[HTML]{B7E1CD}95.43 / \textbf{99.57} & \cellcolor[HTML]{B7E1CD}92.97 / \textbf{99.30} & \cellcolor[HTML]{B7E1CD}96.17 / \textbf{99.73} & \cellcolor[HTML]{B7E1CD}94.89 / \textbf{99.67} & \cellcolor[HTML]{B7E1CD}96.75 / \textbf{99.70} & \cellcolor[HTML]{B7E1CD}97.24 / \textbf{99.86} & \cellcolor[HTML]{B7E1CD}97.72 / \textbf{99.89} \\
\midrule
\end{tabularx}
\setlength{\tabcolsep}{3pt}
\begin{tabularx}{\linewidth}{cc@{\hspace{3.4ex}}c@{\hspace{3.4ex}}c@{\hspace{3.4ex}}c@{\hspace{3.4ex}}c@{\hspace{3.4ex}}c@{\hspace{3.4ex}}c@{\hspace{3.4ex}}c@{\hspace{3.4ex}}c@{\hspace{3.4ex}}c}
Levels   & Snow    & Frost       & Fog    & Brightness & Spatter & Contrast & Elastric & Pixelate & Jpeg  & Saturate\\
\midrule
1 & \cellcolor[HTML]{B7E1CD}87.03 / \textbf{98.95} & \cellcolor[HTML]{B7E1CD}86.85 / \textbf{98.94} & \cellcolor[HTML]{B7E1CD}89.58 / \textbf{99.27} & \cellcolor[HTML]{B7E1CD}84.10 / \textbf{98.48} & \cellcolor[HTML]{B7E1CD}86.45 / \textbf{98.92} & \cellcolor[HTML]{B7E1CD}91.08 / \textbf{99.52} & \cellcolor[HTML]{B7E1CD}91.93 / \textbf{99.68} & \cellcolor[HTML]{B7E1CD}86.28 / \textbf{98.98} & \cellcolor[HTML]{B7E1CD}87.95 / \textbf{99.37} & \cellcolor[HTML]{B7E1CD}87.15 / \textbf{99.22} \\
2 & \cellcolor[HTML]{B7E1CD}90.80 / \textbf{99.21} & \cellcolor[HTML]{B7E1CD}88.08 / \textbf{99.03} & \cellcolor[HTML]{B7E1CD}94.23 / \textbf{99.74} & \cellcolor[HTML]{B7E1CD}83.78 / \textbf{98.43} & \cellcolor[HTML]{B7E1CD}88.80 / \textbf{99.14} & \cellcolor[HTML]{B7E1CD}97.58 / \textbf{99.88} & \cellcolor[HTML]{B7E1CD}92.45 / \textbf{99.65} & \cellcolor[HTML]{B7E1CD}88.48 / \textbf{99.09} & \cellcolor[HTML]{B7E1CD}88.95 / \textbf{99.52} & \cellcolor[HTML]{B7E1CD}87.28 / \textbf{99.22} \\
3 & \cellcolor[HTML]{B7E1CD}88.15 / \textbf{99.17} & \cellcolor[HTML]{B7E1CD}92.23 / \textbf{99.35} & \cellcolor[HTML]{B7E1CD}96.60 / \textbf{99.80} & \cellcolor[HTML]{B7E1CD}83.45 / \textbf{98.58} & \cellcolor[HTML]{B7E1CD}89.33 / \textbf{99.21} & \cellcolor[HTML]{B7E1CD}99.00 / \textbf{99.94} & \cellcolor[HTML]{B7E1CD}95.28 / \textbf{99.78} & \cellcolor[HTML]{B7E1CD}89.38 / \textbf{99.22} & \cellcolor[HTML]{B7E1CD}89.35 / \textbf{99.54} & \cellcolor[HTML]{B7E1CD}83.18 / \textbf{98.19} \\
4 & \cellcolor[HTML]{B7E1CD}88.18 / \textbf{98.92} & \cellcolor[HTML]{B7E1CD}92.68 / \textbf{99.51} & \cellcolor[HTML]{B7E1CD}98.00 / \textbf{99.86} & \cellcolor[HTML]{B7E1CD}83.55 / \textbf{98.51} & \cellcolor[HTML]{B7E1CD}89.43 / \textbf{98.75} & \cellcolor[HTML]{B7E1CD}99.05 / \textbf{99.95} & \cellcolor[HTML]{B7E1CD}95.55 / \textbf{99.85} & \cellcolor[HTML]{B7E1CD}92.25 / \textbf{99.48} & \cellcolor[HTML]{B7E1CD}88.60 / \textbf{99.68} & \cellcolor[HTML]{B7E1CD}82.93 / \textbf{98.08} \\
5 & \cellcolor[HTML]{B7E1CD}89.20 / \textbf{99.17} & \cellcolor[HTML]{B7E1CD}94.93 / \textbf{99.75} & \cellcolor[HTML]{B7E1CD}99.38 / \textbf{99.95} & \cellcolor[HTML]{B7E1CD}83.83 / \textbf{98.57} & \cellcolor[HTML]{B7E1CD}92.73 / \textbf{99.31} & \cellcolor[HTML]{B7E1CD}99.38 / \textbf{99.99} & \cellcolor[HTML]{B7E1CD}95.20 / \textbf{99.75} & \cellcolor[HTML]{B7E1CD}95.05 / \textbf{99.52} & \cellcolor[HTML]{B7E1CD}89.43 / \textbf{99.75} & \cellcolor[HTML]{B7E1CD}83.70 / \textbf{98.27} \\
\midrule
Average & \cellcolor[HTML]{B7E1CD}88.67 / \textbf{99.08} & \cellcolor[HTML]{B7E1CD}90.95 / \textbf{99.32} & \cellcolor[HTML]{B7E1CD}95.56 / \textbf{99.72} & \cellcolor[HTML]{B7E1CD}83.74 / \textbf{98.51} & \cellcolor[HTML]{B7E1CD}89.35 / \textbf{99.06} & \cellcolor[HTML]{B7E1CD}97.22 / \textbf{99.86} & \cellcolor[HTML]{B7E1CD}94.08 / \textbf{99.74} & \cellcolor[HTML]{B7E1CD}90.29 / \textbf{99.26} & \cellcolor[HTML]{B7E1CD}88.86 / \textbf{99.57} & \cellcolor[HTML]{B7E1CD}84.85 / \textbf{98.59} \\
\bottomrule
\end{tabularx}
\end{table*}

Along with known attack detection experiments, we show our approach's unknown attack detection performance against the Mahalanobis method, which showed the second-best performance following ours.
The detectors are trained on attack images of BIM and PGD (denoted as ``seen'') and evaluated on unknown attacks.
The detection performance on the known attack is also included but shown in parentheses for reference.
We observe a similar trend to the known attack detection, where the Mahalanobis method shows saturated performance on the attacks for which the hyperparameters are tuned.
On the contrary, our approach shows more consistent performance across all unknown attacks except Semantic attack with an average of over 95\% AUROC and outperforms the compared method in 18 out of 28 cases.
Our approach can effectively characterize adversarial perturbations even for unknown attacks.

%%%%%%%%%%%%%%%%%%%%%%%%%%%%%%%%%%%%%%%%%%%%%%%%%%%%%%%%%%%%%%%%%%%%%%%%
\subsection{Corrupted Input Detection}\label{ssec:corrupted}

In addition to the widely accepted OOD and adversarial detection setup, we consider corrupted inputs as another type of anomaly.
Deployed in the real world, neural networks are known to suffer from imperfect samples due to the data acquisition process and environmental factors, such as motion blur or weather conditions~\cite{temel2018cureor, temel2019multifarious, hendrycks2019robustness}. 
We use image classification datasets designed to benchmark the robustness of neural networks under realistic challenging conditions. 
CIFAR-10-C~\cite{hendrycks2019robustness} consists of 19 diverse corruption types in four categories, including noise, blur, weather, and digital, at five different severity levels that are applied to test images of CIFAR-10 dataset. 
CURE-TSR~\cite{temel2017curetsr} is a traffic sign recognition dataset that includes real-world and simulated challenging conditions of 12 types and five severity levels. 
For each dataset, a ResNet model is trained on corruption-free images, and the gradients are collected from pristine images in the test sets and their corrupted versions. 
We utilize the Mahalanobis method~\cite{lee2018mahalanobis} for comparison because it showed the best performance among all the compared methods for OOD detection and adversarial detection.

\begin{table*}[t]
\captionsetup{width=\linewidth}
\caption{Classification accuracy for models used in OOD and adversarial detection (\%). For OOD detection, we report the accuracy on test sets of the in-distribution datasets. For adversarial detection, the model trained on clean CIFAR-10 is tested on the adversarial attack images generated using the test set. We also include the accuracy of models used in Section~\ref{sssec:data_dependent}.}
\label{table:classifier_accuracy}
\centering
\small
\begin{tabular}{cccccccccccc}
\toprule
\multirow{2}{*}{\begin{tabular}[c]{@{}c@{}}Network\\ Architecture\end{tabular}} & \multicolumn{2}{c}{OOD} & \multicolumn{7}{c}{Adversarial Attacks on CIFAR-10} & \multicolumn{2}{c}{(OOD)} \\
\cmidrule(l){2-3} \cmidrule(l){4-10} \cmidrule(l){11-12}
&SVHN & CIFAR-10 & \hspace{1mm} FGSM & BIM & C\&W & PGD & IterLL & Semantic & AutoAttack  & MNIST & ImageNet \\
 \midrule
ResNet & 96.65 & 93.84 & \hspace{1mm} 68.17 & 22.37 & 5.38 & 27.12 & 44.31 & 66.1 & 0.06 & 98.24 & 32.06\\
DenseNet & 96.52 & 94.34 & \hspace{1mm} 53.21 & 7.29 & 4.5 & 8.67 & 17.88 & 68.9 & 0.02 & - & - \\
\bottomrule
\end{tabular}
\end{table*}

In the experiments, we observed that the AUROC scores are highly saturated for both methods in many cases, particularly for CIFAR-10-C dataset, which calls for a more comprehensive comparison.
To better facilitate the performance comparison, we employ \textit{corrected repeated $k$-fold cross-validated (CV) paired t-test}~\cite{bouckaert2004evaluating} as a measure of statistical significance.
Paired t-test is a statistical test for comparing two different learning schemes, $A$ and $B$, based on a number of observations; in this case, predictive accuracy $a$ and $b$.
To improve the stability of the test, Nadeau and Bengio~\cite{nadeau1999inference} proposed a corrected variant for the CV setup, and Bouckaert and Frank~\cite{bouckaert2004evaluating} extended it to a repeated CV setup.
For the corrected repeated $k$-fold CV test, each model is trained using $k$-fold cross-validation sets and the process is repeated $r$ times ($r > 1$), resulting in $a_{ij}$ and $b_{ij}$ for each fold $i, 1 \leq i \leq k$ and each run $j, 1 \leq j \leq r$.
Let $x_{ij}$ be the observed difference $x_{ij} = a_{ij} - b_{ij}$, and $m$ and $\sigma^2$ be the estimates for the mean $m = \frac{1}{kr} \sum_{i=1}^k \sum_{j=1}^r x_{ij}$ and variance $\sigma^2 = \frac{1}{kr - 1} \sum_{i=1}^k \sum_{j=1}^r (x_{ij} - m)^2$.
The test statistic $t$ is computed as following:
\begin{equation}
t = \frac{m}{ \sqrt{ ( \frac{1}{kr} + \frac{n_2}{n_1} ) \sigma^2 }}
\end{equation}
where $n_1$ and $n_2$ are the numbers of training and testing instances for the variance estimate correction, respectively.
$t$ is compared to a threshold value based on the significance level $p$ to determine whether the performance of the two models is significantly different (\ie, statistically significant, SS).
In our experiments, we use $k=5$, $r=2$, and $p=0.05$.

We report the detection performance on CURE-TSR in Table~\ref{table:corrupted_curetsr} and CIFAR-10-C in Table~\ref{table:corrupted_cifar10c}.
As the statistical test is conducted using model predictions, we report the performance in terms of detection accuracy rather than AUROC.
We highlight the SS detection accuracy in green and red, where green indicates that our method outperforms the Mahalanobis method, and red shows the opposite case.
Both methods show higher variability in performance across different severity levels of all corruption types for CURE-TSR compared to the relatively saturated results for CIFAR-10-C. 
The proposed method outperforms the compared method in all 95 cases for CIFAR-10-C and 30 out of 37 SS cases (60 overall cases) for CURE-TSR. 
The other 23 cases of CURE-TSR show no statistically significant differences.
Our approach also shows statistically superior performance in 29 out of 31 cases in the average detection accuracy across all severity levels of each corruption type in both datasets.
The remaining two cases show no statistical difference.
Note the superior SS performance of our approach, even at lower severity levels for both datasets.
This shows that our gradient-based representations can characterize corruption more effectively, even at a subtle degree. 
Overall, we highlight that our approach outperforms the activation-based approach in terms of SS results: 94.7\% among the corruption-level-specific performances and 100\% among the corruption-wise average performances.

%%%%%%%%%%%%%%%%%%%%%%%%%%%%%%%%%%%%%%%%%%%%%%%%%%%%%%%%%%%%%%%%
\subsection{Implementation Details}\label{ssec:implementation}

\vspace{1mm}\noindent\textbf{Classification Networks.} 
For the image classification task, we employ two state-of-the-art neural network architectures, which are then utilized to generate gradient-based representations with \textit{confounding labels}: ResNet~\cite{he2016resnet} and DenseNet~\cite{huang2017densenet}. 
We use ResNet with 18 layers and DenseNet with 100 layers, growth rate $k=12$, and dropout rate of $0$. 
Both networks with no pre-training are trained to minimize the cross-entropy loss with SGD optimization (Nesterov momentum factor of $0.9$ and weight decay of \num{5e-4}. 
A learning rate scheduler is implemented to decay the learning rate by a factor of $0.1$ at 50\% and 75\% of the total number of training epochs. 
The models are trained for 300 epochs with a starting learning rate of $0.1$ and batch size of 64. 
The classification accuracy on test sets for the models used in the paper is listed in Table~\ref{table:classifier_accuracy}.

\vspace{1mm}\noindent\textbf{Detection Networks.} 
For anomalous input detection tasks, we utilize a simple multilayer perceptron (MLP) architecture of two fully-connected layers with ReLU non-linearity and dropout following the first fully-connected layer and Sigmoid after the second layer. 
Given an input of gradient-based representations with dimension $d$, the fully-connected layers have dimensions of $d \times 40$ and $40 \times 1$. 
The detectors are trained to minimize the binary cross-entropy loss using the Adam optimizer and a learning rate of \num{1e-3} for 30 epochs. 

\vspace{1mm}\noindent\textbf{Adversarial Attack Hyperparameters.}
To generate adversarial attack images, we borrow publicly available code bases\footnote{\url{https://github.com/Harry24k/adversarial-attacks-pytorch}, \url{https://github.com/cleverhans-lab/cleverhans}} and utilize the hyperparameter values listed in Table~\ref{table:adv_hyperparams}.

\begin{itemize}[leftmargin=9mm,itemsep=0.8mm,topsep=2mm]
    \item $\epsilon$ : maximum magnitude of perturbation 
    \item $\alpha$ : step size in each iteration for iterative attacks
    \item $c$ : box constraint weight for optimization
    \item $k$ : confidence threshold for misclassification
    \item \textit{steps} : number of iterations
\end{itemize}

\begin{table}[ht]
\captionsetup{width=\linewidth}
\caption{Hyperparameters for adversarial attacks.}
\label{table:adv_hyperparams}
\small
\centering
\begin{tabular}{ccc@{\hspace{6mm}}c@{\hspace{6mm}}c@{\hspace{6mm}}c}
\toprule
 & $\epsilon$ & $\alpha$ & $c$ & $k$ & steps \\
 \midrule
FGSM & 0.03 & - & - & - & - \\
BIM & 0.03 & 0.008 & - & - & 100 \\
C\&W & - & - & 1 & 0 & 1000 \\
PGD & 0.03 & 0.008 & - & - & 10 \\
IterLL & 0.05 & 0.005 & - & - & 15 \\
AutoAttack & 0.03 & - & - & - & 100 \\
\bottomrule
\end{tabular}
\end{table}

%%%%%%%%%%%%%%%%%%%%%%%%%%%%%%%%%%%%%%%%%%%%%%%%%%%%%%%%%%%%%%%%
\subsection{Ablation Studies}\label{ssec:ablation}

Our approach to probing the \textit{purview} of neural networks is free of hyperparameters.
It involves some parameters that can be adjusted, but they remain fixed for the experiments shown throughout the paper: 1) the design of \textit{confounding labels} and 2) the anomalous input detector architecture and training.
This section provides detailed ablation studies on these two factors to further validate that our method does not require hyperparameter tuning.

\begin{table*}[t]
\caption{Detection results using NR \textit{confounding labels} for ablation studies in AUROC.}
\label{table:ablation_nr}
\centering
\begin{small}
\begin{tabular}{cccccccccccc}
\toprule
\multicolumn{2}{c}{\multirow{2}{*}{\begin{tabular}[c]{@{}c@{}}Confounding Label\\ Designs\end{tabular}}} & \multicolumn{6}{c}{Adversarial} & \multicolumn{3}{c}{Out-of-Distribution} & \hspace{1mm}\footnotesize{(OOD)}\\
\cmidrule(l){3-8} \cmidrule(l){9-11} \cmidrule(l){12-12}
\multicolumn{2}{c}{} & \hspace{1mm}FGSM & BIM & C\&W & PGD & IterLL & Semantic\hspace{-1mm} & \hspace{1mm}SVHN & ImageNet & LSUN & \hspace{1mm}STL10  \\
\midrule
\multicolumn{2}{c}{All-hot label} 
 & 93.45 & \textbf{96.19} & 97.07 & \textbf{95.82} & \textbf{98.17} & 90.15 & 
 \textbf{99.90} & \textbf{96.88} & \textbf{99.90} & \textbf{74.94} \\
\midrule
\multirow{5}{*}{\begin{tabular}[c]{@{}c@{}}Top-k\\ Prediction\end{tabular}} 
 & 1 & 89.47 & 88.82 & 97.62 & 93.15 & 94.94 & 88.82 & 99.84 & 95.09 & 99.83 & 70.97 \\
 & 2 & 88.98 & 94.29 & 97.64 & 93.31 & 95.09 & 88.61 & 99.83 & 94.89 & 99.87 & 70.37 \\
 & 3 & 89.21 & 94.34 & 97.49 & 93.48 & 94.88 & 89.08 & 99.84 & 94.94 & 99.83 & 70.64 \\
 & 4 & 89.55 & 93.90 & 97.64 & 93.36 & 94.94 & 88.86 & 99.84 & 95.22 & 99.82 & 70.70  \\
 & 5 & 89.46 & 94.40 & 97.63 & 93.49 & 95.13 & 88.43 & 99.82 & 95.25 & 99.81 & 70.46 \\
\midrule
\multirow{2}{*}{Taxanomy}
 & Animal & \textbf{93.65} & 95.61 & 97.16 & 95.00 & 97.33 & 90.68 & 99.82 & 92.09 & 99.81 & 72.88 \\
 & Vehicle &  93.19 & 95.33 & 96.43 & 94.59 & 96.82 & \textbf{90.89} & 99.87 & 92.07 & 99.88 & 72.44 \\
\midrule
% \multirow{2}{*}{(STL-10 only)} & Common & \multicolumn{6}{c}{\multirow{2}{*}{-}} & \multicolumn{3}{c}{\multirow{2}{*}{-}} & 73.64 \\
%  & Unique & \multicolumn{6}{c}{} & \multicolumn{3}{c}{} & \textbf{75.11}\\
% \midrule
\multicolumn{2}{c}{Max Logit Value} & 91.02 & 95.53 & \textbf{97.67} &	94.68 &	96.73 &	89.49 &	99.85 & 95.85 & 99.89 &	71.67 \\
\bottomrule
\end{tabular}
\vspace{-2mm}
\end{small}
\end{table*}

\begin{table}[t]
\captionsetup{width=\linewidth}
\caption{Detection results using FR \textit{confounding labels} in comparison with NR all-hot encoding in AUROC.}
\label{table:ablation_fr}
\centering
\begin{small}
\begin{tabular}{ccccccc}
\toprule
\multicolumn{2}{c}{ \multirow{2}{*}{\begin{tabular}[c]{@{}c@{}}Confounding \\ Designs\end{tabular}}
} & \multicolumn{4}{c}{Adversarial} & \footnotesize{(OOD)}\\
\cmidrule(l){3-6} \cmidrule(l){7-7}
 &  &  FGSM & BIM & C\&W & PGD &  STL10 \\
\midrule
NR & All-hot &  \textbf{93.45} & 96.19 & 97.07 & 95.82 &  74.94 \\
\midrule
\multirow{3}{*}{FR} 
& Target &  92.75 & \textbf{96.78} & \textbf{98.17} & \textbf{96.31} &  - \\
\cmidrule(l){2-7}
& Common & \multicolumn{4}{c}{\multirow{2}{*}{-}} &  73.64 \\
& Unique & \multicolumn{4}{c}{} &  \textbf{75.11} \\
 \bottomrule
\end{tabular}
\vspace{-3mm}
\end{small}
\end{table}

% \begin{sidewaystable*}
\begin{table*}[t]
\captionsetup{width=\linewidth}
\caption{Ablation study on the hyperparameters for the detector network architectures and training. The hyperparameter values used for experiments in Section~\ref{sec:experiments} are underlined for reference.}
\label{table:detector}
\centering
\begin{small}
\begin{tabular}{ccccccccccc}
\toprule
\multicolumn{2}{c}{\multirow{2}{*}{Hyperparameter}} & \multicolumn{3}{c}{SVHN} & \multicolumn{3}{c}{ImageNet} & \multicolumn{3}{c}{LSUN} \\
\cmidrule(l){3-5} \cmidrule(l){6-8} \cmidrule(l){9-11}
\multicolumn{2}{c}{} & Accuracy & AUROC & AUPR & Accuracy & AUROC & AUPR & Accuracy & AUROC & AUPR \\
\midrule
\multirow{3}{*}{Layers} 
 & \underline{2} & 98.44 & 99.90 & 99.99 & 90.97 & 96.88 & 97.20 & 98.59 & 99.90 & 99.91 \\
 & 3 & 98.90 & \textbf{99.95} & \textbf{99.99} & \textbf{91.75} & \textbf{97.16} & \textbf{97.28} & \textbf{99.22} & \textbf{99.97} & \textbf{99.97} \\
 & 4 & \textbf{99.01} & \textbf{99.95} & \textbf{99.99} & 91.40 & 97.12 & 97.24 & 99.14 & 99.95 & 99.95 \\
 \midrule
\multirow{8}{*}{Neurons} 
 & 10 & 98.08 & 99.84 & 99.98 & 90.73 & 96.54 & 96.78 & 98.04 & 99.80 & \textbf{99.81} \\
 & 15 & 98.05 & 99.83 & 99.98 & 90.84 & 96.74 & 96.92 & 98.40 & 99.86 & 99.86 \\
 & 20 & 98.21 & 99.86 & 99.98 & 90.92 & 96.73 & 96.92 & 98.57 & \textbf{99.90} & \textbf{99.91} \\
 & 25 & 98.33 & 99.87 & \textbf{99.99} & 91.09 & 96.87 & 97.08 & \textbf{98.66} & \textbf{99.90} & \textbf{99.91} \\
 & 30 & 98.25 & 99.87 & \textbf{99.99} & 90.92 & 96.78 & 97.15 & 98.61 & \textbf{99.90} & \textbf{99.91} \\
 & 35 & 98.31 & 99.88 & \textbf{99.99} & 90.96 & 96.82 & 96.90 & 98.60 & 99.89 & 99.90 \\
 & \underline{40} & \textbf{98.44} & \textbf{99.90} & \textbf{99.99} & 90.97 & 96.88 & 97.20 & 98.59 & \textbf{99.90} & \textbf{99.91} \\
 & 45 & 98.37 & \textbf{99.90} & \textbf{99.99} & \textbf{91.23} & \textbf{97.00} & \textbf{97.42} & 98.62 & \textbf{99.90} & 99.90 \\
 \midrule
\multirow{7}{*}{Epochs} 
 & 10 & 97.45 & 99.76 & 99.97 & 91.07 & 96.74 & 97.09 & 97.64 & 99.78 & 99.78 \\
 & 15 & 98.08 & 99.84 & 99.98 & 90.78 & 96.69 & 96.96 & 97.99 & 99.81 & 99.82 \\
 & 20 & 98.06 & 99.87 & \textbf{99.99} & 90.82 & 96.70 & 96.80 & 98.57 & 99.88 & 99.88 \\
 & 25 & 98.46 & 99.89 & \textbf{99.99} & \textbf{91.09} & 96.79 & 97.00 & 98.69 & \textbf{99.92} & \textbf{99.92} \\
 & \underline{30} & 98.44 & \textbf{99.90} & \textbf{99.99} & 90.97 & \textbf{96.88} & \textbf{97.20} & 98.59 & 99.90 & 99.91 \\
  & 35 & \textbf{98.57} &	\textbf{99.90} &	\textbf{99.99} &	90.88 &	96.81 &	97.15 &	\textbf{98.82} &	99.91 &	\textbf{99.92}\\
 \midrule
\multirow{6}{*}{\begin{tabular}[c]{@{}c@{}}Learning\\ Rates\end{tabular}} 
 & 0.1 & 98.73 & 99.90 & \textbf{99.99} & 91.15 & 96.76 & 96.84 & 98.89 & 99.89 & 99.87 \\
 & 0.05 & 98.80 & 99.92 & \textbf{99.99} & 91.33 & 96.95 & 97.31 & 98.98 & 99.93 & 99.94 \\
 & 0.01 & \textbf{98.97} & \textbf{99.94} & \textbf{99.99} & \textbf{91.54} & 97.04 & \textbf{97.35} & \textbf{99.08} & \textbf{99.96} & \textbf{99.96} \\
 & 0.005 & 98.87 & 99.93 & \textbf{99.99} & 91.47 & \textbf{97.11} & \textbf{97.35} & 99.06 & \textbf{99.96} & \textbf{99.96} \\
 & \underline{0.001} & 98.44 & 99.90 & \textbf{99.99} & 90.97 & 96.88 & 97.20 & 98.59 & 99.90 & 99.91 \\
 & 0.0005 & 97.96 & 99.83 & 99.98 & 90.91 & 96.77 & 97.13 & 98.17 & 99.87 & 99.88 \\
\bottomrule
\end{tabular}
\end{small}\vspace{-2mm}
\end{table*}
% \end{sidewaystable*}

%%%%%%%%%%%%%%%%%%%%%%%%%%%%%%%%%%%%%%%%%%%%%%%%%%%%%%%%%%%%%%%%
\subsubsection{Confounding Label Designs}\label{sssec:ablation_cflabel}

We introduced \textit{confounding labels} as a methodology to elicit a gradient response without relying on any information regarding the inputs given during inference.
The \textit{confounding labels} are devised by combining the one-hot encodings of multiple classes, and we utilize the default choice of the \textit{confounding label} where the one-hot encodings of all classes are combined (\ie, an all-hot label) throughout the experiment section.
In this section, we present ablation studies on different designs of \textit{confounding labels}, as opposed to the choice of all-hot labels in generating gradients.
We establish two types of \textit{confounding labels} based on reference to information regarding inputs during inference: no-reference (NR) and full-reference (FR) designs.
The terminologies are borrowed from the field of image quality assessment~\cite{temel2016unique}, where the reference to clean images is a critical factor in evaluating the quality of corrupted images.
Our goal is to probe the \textit{purview} of neural networks during inference, where we generally do not have access to any information regarding the given inputs.
Nonetheless, we consider both options to explore the effect of \textit{confounding label} designs and show the most viable options in each case in the OOD and adversarial detection setups.
Following the experimental setup presented in Sections \ref{ssec:ood} and \ref{ssec:adversarial}, we utilize a ResNet classifier trained on CIFAR-10.
For adversarial detection, we employ FGSM, BIM, C\&W, PGD, IterLL, and Semantic attacks.
For OOD detection, we utilize the originally utilized SVHN, resized LSUN, resized ImageNet, and an additional dataset of STL-10~\cite{coates2011stl10}, given the shared classes with CIFAR-10, which allow for unique insights.

\vspace{1mm}\noindent\textbf{No-Reference (NR) Labels.} 
To construct NR \textit{confounding labels}, we only utilize the information that is safely assumed to be available at inference time: trained classes of the classifier and model outputs in response to given inputs during inference.
We formulate three types of NR labels based on 1) top-$k$ predictions, 2) taxonomy of trained classes, and 3) maximum logit values.
The top-$k$ prediction-based \textit{confounding labels} are devised by combining the one-hot encodings of the top $1$ through $k$ classes based on the predicted class probabilities.
The class taxonomy-based labels are constructed based on the two superclasses of CIFAR-10: animals (six subclasses: bird, cat, deer, dog, frog, and horse) and vehicles (four subclasses: airplane, car, ship, and truck).
For each superclass, the labels are implemented by combining the one-hot encodings of the subclasses.
The logit-based label design is based on the literature introduced in Section~\ref{ssec:gradient_features}, where the maximum logit value is directly backpropagated in generating gradients instead of loss values between the model outputs and some labels.

We employ the three types of NR labels in gradient generation for adversarial and OOD detection and report the detection performance in AUROC in Table~\ref{table:ablation_nr}.
Overall, the all-hot label performs best in 7 out of 10 cases.
For adversarial detection, all-hot label performance is followed by taxonomy-based labels and maximum logit values.
Top-$k$ prediction-based labels are the least effective in generating gradient-based representations that capture perturbations.
This is because of the nature of adversarial attacks, specifically designed to deceive the classifiers and result in perturbed probability distributions.
Consequently, the top-$k$ predictions are unreliable for eliciting meaningful gradient responses.
On the other hand, the trained class-based labels can generate gradients for specific semantic categories, which capture adversarial perturbations more effectively.
For OOD detection, the results exhibit less variation, except for the inferior performance of taxonomy-based labels for ImageNet and superior performance for STL-10.
Contrary to adversarial attacks, OOD samples are not intentionally devised for incorrect classification but are drawn far from the training data distribution.
The trained classes of CIFAR-10 may be insufficient for the exponentially larger number of classes represented in the ImageNet dataset, leading to an inferior performance.
Conversely, the shared classes between STL-10 and CIFAR-10 led to more relevant gradient responses, and thus, superior detection performance.
Overall, the all-hot \textit{confounding label} is the most viable design for generating gradient-based representations that can distinguish adversarial and OOD samples from clean and in-distribution data, with no reference to information regarding the data samples.

\vspace{1mm}\noindent\textbf{Full-Reference (FR) Labels.} 
When there is no available information concerning the given inputs during inference, the all-hot \textit{confounding label} is proven to be the best option.
However, if we have access to information regarding the inputs, we can devise better \textit{confounding label} designs.
For adversarial detection, we consider targeted adversarial attacks where the perturbation is added to the images so that they can be classified as specific target classes.
The target classes can then be used to formulate the \textit{confounding labels} using their one-hot encodings.
For OOD detection, we utilize STL-10 dataset for its nine common classes with CIFAR-10 and one unique class.
A common class-based \textit{confounding label} can be constructed by combining the one-hot encodings of the nine common classes and the unique class-based label with the unique class' one-hot encoding.

The detection results obtained using FR labels are shown in Table~\ref{table:ablation_fr}.
For adversarial detection, we utilize four attack types (FGSM, BIM, C\&W, and PGD) and present the average results across all target classes.
The target-class-based \textit{confounding labels} perform better than the all-hot labels for three out of four adversarial attacks.
For OOD detection with STL-10, the unique class-based label performs better than the all-hot label, and the common class-based labels perform worse than the all-hot label.
The unique class-based \textit{confounding label} can elicit gradient responses with respect to the features of the unique class of CIFAR-10, which are absent in STL-10.
As a result, the gradient distributions exhibit a more significant gap between the datasets and lead to a better OOD detection performance.
Overall, the knowledge about inputs given during inference has proven useful in designing \textit{confounding labels} to generate gradient responses that can better characterize the anomaly in inputs, leading to improved adversarial and OOD detection performances compared to NR label designs.

%%%%%%%%%%%%%%%%%%%%%%%%%%%%%%%%%%%%%%%%%%%%%%%%%%%%%%%%%%%%%%%%
\subsubsection{Detector Designs and Training}\label{sssec:ablation_detector}

We utilize a simple multilayer perceptron (MLP) as an anomalous input detector using gradients.
This section provides ablation studies on the detector network design and its training.
We consider the ablation settings for the following hyperparameters:
\vspace{1mm}
\begin{itemize}[leftmargin=5mm,itemsep=0.8mm,topsep=1mm]
    \item Layers: number of layers for the MLP 
    \item Neurons: number of neurons in each layer
    \item Epochs: number of passes through the dataset in training
    \item Learning rates: step size at each iteration in optimization
\end{itemize}\vspace{1mm}
Following Section~\ref{ssec:ood}, we utilize the OOD detection setup with a ResNet classifier and CIFAR-10 dataset as in-distribution.
The detection performances using various hyperparameter values are reported in Table~\ref{table:detector}, measured in detection accuracy, AUROC, and AUPR, using 5-fold cross-validation.
The values used for the experiments in Sections~\ref{ssec:ood} through \ref{ssec:corrupted} are referred to as the default values and are underlined for reference.
For the ablation of each hyperparameter, all other hyperparameters are held constant at the default values, and only the specific hyperparameter in question is changed.

We observe that varying the hyperparameter values has little effect on the detection performance for all hyperparameters.
The maximum AUROC gap observed for SVHN and LSUN is only 0.14\% when varying the number of epochs and 0.46\% for ImageNet with a different number of neurons.
The effect of different hyperparameter values only leads to an average AUROC gap of 0.09\% for SVHN, 0.1\% for LSUN, and 0.32\% for ImageNet.
Note that the default values for the hyperparameters do not always lead to the best performance.
This ablation study further highlights the simplicity of our approach with no hyperparameters involved in detector design and training as well as generating gradients to effectively characterize the anomaly in inputs.
The gradient-based representations generated with \textit{confounding labels} have proven useful in differentiating anomalous inputs from the perspective of models without relying on a sophisticated detector.

\section{Conclusion}
In this paper, we propose to examine the data-dependent \textit{effective capacity} of neural networks to probe their \textit{purview}.
We define \textit{purview} of a model as the capacity required to characterize given samples during inference, in addition to the capacity defined by its training data.
Inspired by the utility of gradients used in model training, we utilize gradients to measure the amount of change required for a model to characterize inputs more accurately during inference.
To facilitate gradient generation during inference, we introduce \textit{confounding labels} that can be formulated with no information regarding the given inputs.
We validate the effectiveness of gradients generated with \textit{confounding labels} in capturing anomalies in inputs for detecting OOD, adversarial, and corrupted inputs.
Finally, we provide extensive ablation studies for the design of \textit{confounding labels} and hyperparameter settings.

\bibliographystyle{IEEEtran}
\bibliography{bib.bib}

% Generated by IEEEtran.bst, version: 1.14 (2015/08/26)
\begin{thebibliography}{10}
\providecommand{\url}[1]{#1}
\csname url@samestyle\endcsname
\providecommand{\newblock}{\relax}
\providecommand{\bibinfo}[2]{#2}
\providecommand{\BIBentrySTDinterwordspacing}{\spaceskip=0pt\relax}
\providecommand{\BIBentryALTinterwordstretchfactor}{4}
\providecommand{\BIBentryALTinterwordspacing}{\spaceskip=\fontdimen2\font plus
\BIBentryALTinterwordstretchfactor\fontdimen3\font minus
  \fontdimen4\font\relax}
\providecommand{\BIBforeignlanguage}[2]{{%
\expandafter\ifx\csname l@#1\endcsname\relax
\typeout{** WARNING: IEEEtran.bst: No hyphenation pattern has been}%
\typeout{** loaded for the language `#1'. Using the pattern for}%
\typeout{** the default language instead.}%
\else
\language=\csname l@#1\endcsname
\fi
#2}}
\providecommand{\BIBdecl}{\relax}
\BIBdecl

\bibitem{temel2018cureor}
D.~Temel, J.~Lee, and G.~AlRegib, ``Cure-or: Challenging unreal and real
  environments for object recognition,'' in \emph{IEEE International Conference
  on Machine Learning and Applications}, 2018, pp. 137--144.

\bibitem{hendrycks2019robustness}
D.~Hendrycks and T.~Dietterich, ``Benchmarking neural network robustness to
  common corruptions and perturbations,'' \emph{Proceedings of the
  International Conference on Learning Representations}, 2019.

\bibitem{temel2019multifarious}
D.~Temel, J.~Lee, and G.~AlRegib, ``Object recognition under multifarious
  conditions: A reliability analysis and a feature similarity-based performance
  estimation,'' in \emph{IEEE International Conference on Image Processing},
  2019, pp. 3033--3037.

\bibitem{goodfellow2014adversarial}
I.~Goodfellow, J.~Shlens, and C.~Szegedy, ``Explaining and harnessing
  adversarial examples,'' in \emph{International Conference on Learning
  Representations}, 2015.

\bibitem{guo2017calibration}
C.~Guo, G.~Pleiss, Y.~Sun, and K.~Q. Weinberger, ``On calibration of modern
  neural networks,'' in \emph{Proceedings of the International Conference on
  Machine Learning-Volume 70}.\hskip 1em plus 0.5em minus 0.4em\relax JMLR.
  org, 2017, pp. 1321--1330.

\bibitem{lee2020gradients}
J.~Lee and G.~AlRegib, ``Gradients as a measure of uncertainty in neural
  networks,'' in \emph{IEEE International Conference on Image Processing},
  2020.

\bibitem{lee2022gradient}
J.~Lee, M.~Prabhushankar, and G.~AlRegib, ``Gradient-based adversarial and
  out-of-distribution detection,'' \emph{ICML Workshop on New Frontiers in
  Adversarial Machine Learning}, 2022.

\bibitem{temel2017curetsr}
D.~Temel, G.~Kwon, M.~Prabhushankar, and G.~AlRegib, ``Cure-tsr: Challenging
  unreal and real environments for traffic sign recognition,'' \emph{NeurIPS
  Workshop on Machine Learning for Intelligent Transportation Systems}, 2017.

\bibitem{Logan2022ISBI}
G.~K. Y.~Logan, K.~Kokilepersaud, C.~W. G.~AlRegib, and H.~Yu, ``Multi-modal
  learning using physicians diagnostics for optical coherence tomography
  classification,'' \emph{IEEE International Symposium on Biomedical Imaging},
  2022.

\bibitem{bengio2009learning}
Y.~Bengio \emph{et~al.}, ``Learning deep architectures for ai,''
  \emph{Foundations and trends{\textregistered} in Machine Learning}, vol.~2,
  no.~1, pp. 1--127, 2009.

\bibitem{dauphin2013big}
Y.~N. Dauphin and Y.~Bengio, ``Big neural networks waste capacity,''
  \emph{arXiv preprint arXiv:1301.3583}, 2013.

\bibitem{goodfellow2016deep}
I.~Goodfellow, Y.~Bengio, and A.~Courville, \emph{Deep learning}.\hskip 1em
  plus 0.5em minus 0.4em\relax MIT press, 2016.

\bibitem{vapnik1971uniform}
V.~N. Vapnik and A.~Y. Chervonenkis, ``On the uniform convergence of relative
  frequencies of events to their probabilities,'' in \emph{Theory of
  Probability and Its Applications}, 1971.

\bibitem{neyshabur2014search}
\BIBentryALTinterwordspacing
B.~Neyshabur, R.~Tomioka, and N.~Srebro, ``In search of the real inductive
  bias: On the role of implicit regularization in deep learning.'' in
  \emph{International Conference on Learning Representation (ICLR) Workshop},
  2015. [Online]. Available: \url{http://arxiv.org/abs/1412.6614}
\BIBentrySTDinterwordspacing

\bibitem{zhang2021understanding}
C.~Zhang, S.~Bengio, M.~Hardt, B.~Recht, and O.~Vinyals, ``Understanding deep
  learning (still) requires rethinking generalization,'' \emph{Communications
  of the ACM}, vol.~64, no.~3, pp. 107--115, 2021.

\bibitem{cybenko1989approximation}
G.~Cybenko, ``Approximation by superpositions of a sigmoidal function,''
  \emph{Mathematics of control, signals and systems}, vol.~2, no.~4, pp.
  303--314, 1989.

\bibitem{montufar2014number}
G.~F. Montufar, R.~Pascanu, K.~Cho, and Y.~Bengio, ``On the number of linear
  regions of deep neural networks,'' \emph{Advances in neural information
  processing systems}, vol.~27, 2014.

\bibitem{tokozume2018between}
Y.~Tokozume, Y.~Ushiku, and T.~Harada, ``Between-class learning for image
  classification,'' in \emph{Proceedings of the IEEE Conference on Computer
  Vision and Pattern Recognition}, 2018, pp. 5486--5494.

\bibitem{zhang2017mixup}
H.~Zhang, M.~Cisse, Y.~N. Dauphin, and D.~Lopez-Paz, ``mixup: Beyond empirical
  risk minimization,'' \emph{arXiv preprint arXiv:1710.09412}, 2017.

\bibitem{yun2019cutmix}
S.~Yun, D.~Han, S.~Chun, S.~J. Oh, J.~Choe, and Y.~Yoo, ``Cutmix:
  Regularization strategy to train strong classifiers with localizable
  features,'' in \emph{Proceedings of the IEEE International Conference on
  Computer Vision}, 2019.

\bibitem{durand2019learning}
T.~Durand, N.~Mehrasa, and G.~Mori, ``Learning a deep convnet for multi-label
  classification with partial labels,'' in \emph{Proceedings of the IEEE/CVF
  Conference on Computer Vision and Pattern Recognition}, 2019, pp. 647--657.

\bibitem{duarte2021plm}
K.~Duarte, Y.~Rawat, and M.~Shah, ``Plm: Partial label masking for imbalanced
  multi-label classification,'' in \emph{Proceedings of the IEEE/CVF Conference
  on Computer Vision and Pattern Recognition}, 2021, pp. 2739--2748.

\bibitem{mei2022towards}
S.~Mei, C.~Zhao, B.~Ni, and S.~Yuan, ``Towards bridging sample complexity and
  model capacity,'' in \emph{Proceedings of the AAAI Conference on Artificial
  Intelligence}, 2022.

\bibitem{schmidt2018adversarially}
L.~Schmidt, S.~Santurkar, D.~Tsipras, K.~Talwar, and A.~Madry, ``Adversarially
  robust generalization requires more data,'' \emph{Advances in neural
  information processing systems}, vol.~31, 2018.

\bibitem{li_id_2018_ICLR}
C.~Li, H.~Farkhoor, R.~Liu, and J.~Yosinski, ``Measuring the intrinsic
  dimension of objective landscapes,'' in \emph{International Conference on
  Learning Representations}, 2018.

\bibitem{ansuini2019intrinsic}
A.~Ansuini, A.~Laio, J.~H. Macke, and D.~Zoccolan, ``Intrinsic dimension of
  data representations in deep neural networks,'' \emph{Advances in Neural
  Information Processing Systems}, vol.~32, 2019.

\bibitem{arpit2017closer}
D.~Arpit, S.~Jastrz{\k{e}}bski, N.~Ballas, D.~Krueger, E.~Bengio, M.~S. Kanwal,
  T.~Maharaj, A.~Fischer, A.~Courville, Y.~Bengio \emph{et~al.}, ``A closer
  look at memorization in deep networks,'' in \emph{International conference on
  machine learning}.\hskip 1em plus 0.5em minus 0.4em\relax PMLR, 2017, pp.
  233--242.

\bibitem{hendrycks2016baseline}
D.~Hendrycks and K.~Gimpel, ``A baseline for detecting misclassified and
  out-of-distribution examples in neural networks,'' \emph{Proceedings of the
  International Conference on Learning Representations}, 2017.

\bibitem{liang2018odin}
S.~Liang, Y.~Li, and R.~Srikant, ``Enhancing the reliability of
  out-of-distribution image detection in neural networks,'' in
  \emph{Proceedings of International Conference on Learning Representations},
  2018.

\bibitem{devries2018confidence}
T.~DeVries and G.~W. Taylor, ``Learning confidence for out-of-distribution
  detection in neural networks,'' \emph{arXiv preprint arXiv:1802.04865}, 2018.

\bibitem{ma2018lid}
X.~Ma, B.~Li, Y.~Wang, S.~M. Erfani, S.~Wijewickrema, G.~Schoenebeck, D.~Song,
  M.~E. Houle, and J.~Bailey, ``Characterizing adversarial subspaces using
  local intrinsic dimensionality,'' in \emph{Proceedings of the International
  Conference on Learning Representations}, 2018.

\bibitem{lee2018mahalanobis}
K.~Lee, K.~Lee, H.~Lee, and J.~Shin, ``A simple unified framework for detecting
  out-of-distribution samples and adversarial attacks,'' in \emph{Advances in
  Neural Information Processing Systems}, 2018, pp. 7167--7177.

\bibitem{liu2020energyOOD}
W.~Liu, X.~Wang, J.~Owens, and Y.~Li, ``Energy-based out-of-distribution
  detection,'' in \emph{Advances in Neural Information Processing Systems},
  2020.

\bibitem{rumelhart1986learning}
D.~E. Rumelhart, G.~E. Hinton, and R.~J. Williams, ``Learning representations
  by back-propagating errors,'' \emph{nature}, vol. 323, no. 6088, pp.
  533--536, 1986.

\bibitem{kurakin2017bim_iterll}
A.~Kurakin, I.~Goodfellow, and S.~Bengio, ``Adversarial examples in the
  physical world,'' \emph{International Conference on Learning Representations
  Workshop}, 2017.

\bibitem{madry2017pgd}
A.~Madry, A.~Makelov, L.~Schmidt, D.~Tsipras, and A.~Vladu, ``Towards deep
  learning models resistant to adversarial attacks,'' in \emph{Proceedings of
  the International Conference on Learning Representations}, 2017.

\bibitem{selvaraju2017gradcam}
R.~R. Selvaraju, M.~Cogswell, A.~Das, R.~Vedantam, D.~Parikh, and D.~Batra,
  ``Grad-cam: Visual explanations from deep networks via gradient-based
  localization,'' in \emph{Proceedings of the IEEE international conference on
  computer vision}, 2017, pp. 618--626.

\bibitem{chattopadhay2018grad}
A.~Chattopadhay, A.~Sarkar, P.~Howlader, and V.~N. Balasubramanian,
  ``Grad-cam++: Generalized gradient-based visual explanations for deep
  convolutional networks,'' in \emph{2018 IEEE winter conference on
  applications of computer vision (WACV)}.\hskip 1em plus 0.5em minus
  0.4em\relax IEEE, 2018, pp. 839--847.

\bibitem{alregib2022explanatory}
G.~AlRegib and M.~Prabhushankar, ``Explanatory paradigms in neural networks:
  Towards relevant and contextual explanations,'' \emph{IEEE Signal Processing
  Magazine}, vol.~39, no.~4, pp. 59--72, 2022.

\bibitem{prabhushankar2022introspective}
M.~Prabhushankar and G.~AlRegib, ``Introspective learning: A two-stage approach
  for inference in neural networks,'' \emph{Advances in neural information
  processing systems}, 2022.

\bibitem{kwon2020backpropagated}
G.~Kwon, M.~Prabhushankar, D.~Temel, and G.~AlRegib, ``Backpropagated gradient
  representations for anomaly detection,'' in \emph{Proceedings of the European
  Conference on Computer Vision}, 2020.

\bibitem{oberdiek2018classification}
P.~Oberdiek, M.~Rottmann, and H.~Gottschalk, ``Classification uncertainty of
  deep neural networks based on gradient information,'' in \emph{IAPR Workshop
  on Artificial Neural Networks in Pattern Recognition}, 2018, pp. 113--125.

\bibitem{huang2021gradnorm}
R.~Huang, A.~Geng, and Y.~Li, ``On the importance of gradients for detecting
  distributional shifts in the wild,'' \emph{Advances in Neural Information
  Processing Systems}, vol.~34, pp. 677--689, 2021.

\bibitem{prabhushankar2021extracting}
M.~Prabhushankar and G.~AlRegib, ``Extracting causal visual features for
  limited label classification,'' in \emph{2021 IEEE International Conference
  on Image Processing (ICIP)}.\hskip 1em plus 0.5em minus 0.4em\relax IEEE,
  2021, pp. 3697--3701.

\bibitem{he2016resnet}
K.~He, X.~Zhang, S.~Ren, and J.~Sun, ``Deep residual learning for image
  recognition,'' in \emph{Proceedings of the IEEE Conference on Computer Vision
  and Pattern Recognition}, 2016, pp. 770--778.

\bibitem{lecun1998mnist}
Y.~LeCun, L.~Bottou, Y.~Bengio, and P.~Haffner, ``Gradient-based learning
  applied to document recognition,'' \emph{Proceedings of the IEEE}, vol.~86,
  no.~11, pp. 2278--2324, 1998.

\bibitem{netzer2011svhn}
Y.~Netzer, T.~Wang, A.~Coates, A.~Bissacco, B.~Wu, and A.~Y. Ng, ``Reading
  digits in natural images with unsupervised feature learning,'' \emph{NeurlIPS
  Workshop on Deep Learning and Unsupervised Feature Learning}, 2011.

\bibitem{krizhevsky2009cifar}
A.~Krizhevsky and G.~Hinton, ``Learning multiple layers of features from tiny
  images,'' Tech. Rep., 2009.

\bibitem{yu2015lsun}
F.~Yu, A.~Seff, Y.~Zhang, S.~Song, T.~Funkhouser, and J.~Xiao, ``Lsun:
  Construction of a large-scale image dataset using deep learning with humans
  in the loop,'' \emph{arXiv preprint arXiv:1506.03365}, 2015.

\bibitem{deng2009imagenet}
J.~Deng, W.~Dong, R.~Socher, L.-J. Li, K.~Li, and L.~Fei-Fei, ``Imagenet: A
  large-scale hierarchical image database,'' in \emph{IEEE Conference on
  Computer Vision and Pattern Recognition}, 2009, pp. 248--255.

\bibitem{tack2020csi}
J.~Tack, S.~Mo, J.~Jeong, and J.~Shin, ``Csi: Novelty detection via contrastive
  learning on distributionally shifted instances,'' \emph{Advances in neural
  information processing systems}, vol.~33, pp. 11\,839--11\,852, 2020.

\bibitem{huang2017densenet}
G.~Huang, Z.~Liu, L.~Van Der~Maaten, and K.~Q. Weinberger, ``Densely connected
  convolutional networks,'' in \emph{Proceedings of the IEEE Conference on
  Computer Vision and Pattern Recognition}, 2017, pp. 4700--4708.

\bibitem{carlini2017cw}
N.~Carlini and D.~Wagner, ``Towards evaluating the robustness of neural
  networks,'' in \emph{IEEE Symposium on Security and Privacy}, 2017, pp.
  39--57.

\bibitem{hosseini2017semantic}
H.~Hosseini, B.~Xiao, M.~Jaiswal, and R.~Poovendran, ``On the limitation of
  convolutional neural networks in recognizing negative images,'' in \emph{IEEE
  International Conference on Machine Learning and Applications}, 2017, pp.
  352--358.

\bibitem{croce2020autoattack}
F.~Croce and M.~Hein, ``Reliable evaluation of adversarial robustness with an
  ensemble of diverse parameter-free attacks,'' in \emph{International
  conference on machine learning}.\hskip 1em plus 0.5em minus 0.4em\relax PMLR,
  2020, pp. 2206--2216.

\bibitem{hu2019new}
S.~Hu, T.~Yu, C.~Guo, W.-L. Chao, and K.~Q. Weinberger, ``A new defense against
  adversarial images: Turning a weakness into a strength,'' \emph{Advances in
  Neural Information Processing Systems}, vol.~32, 2019.

\bibitem{bouckaert2004evaluating}
R.~R. Bouckaert and E.~Frank, ``Evaluating the replicability of significance
  tests for comparing learning algorithms,'' in \emph{PAKDD}, vol. 3056.\hskip
  1em plus 0.5em minus 0.4em\relax Springer, 2004, pp. 3--12.

\bibitem{nadeau1999inference}
C.~Nadeau and Y.~Bengio, ``Inference for the generalization error,''
  \emph{Advances in neural information processing systems}, vol.~12, 1999.

\bibitem{temel2016unique}
D.~Temel, M.~Prabhushankar, and G.~AlRegib, ``Unique: Unsupervised image
  quality estimation,'' \emph{IEEE signal processing letters}, vol.~23, no.~10,
  pp. 1414--1418, 2016.

\bibitem{coates2011stl10}
A.~Coates, A.~Ng, and H.~Lee, ``An analysis of single-layer networks in
  unsupervised feature learning,'' in \emph{Proceedings of the International
  Conference on Artificial Intelligence and Statistics}, 2011, pp. 215--223.

\end{thebibliography}

\begin{IEEEbiography}[{\includegraphics[width=1in,height=1.25in,clip,keepaspectratio]{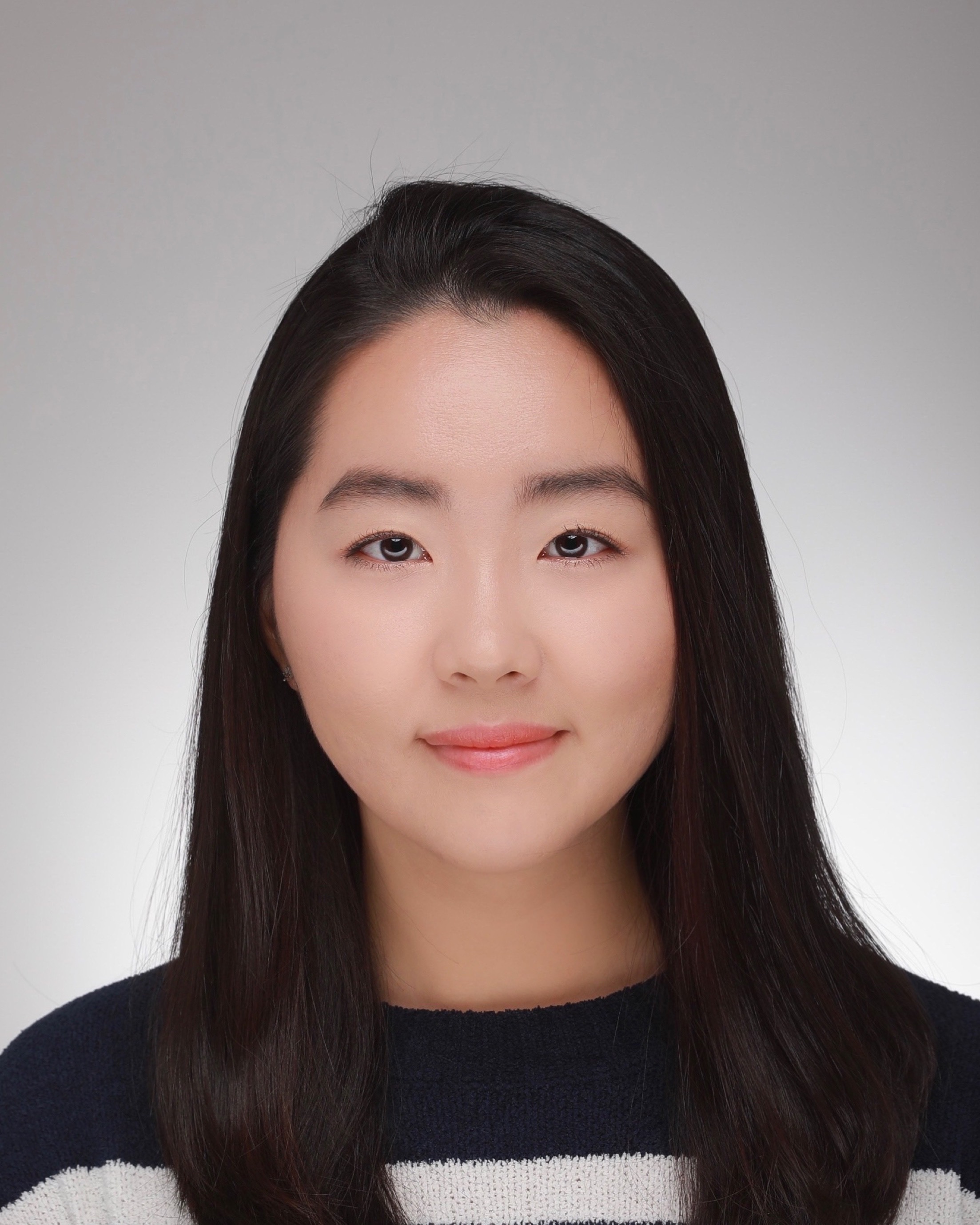}}]{Jinsol Lee} (Member, IEEE) received the B.S. degree in electrical engineering from Purdue University and the M.S. degree in electrical and computer engineering from Georgia Institute of Technology, where she is currently pursuing the Ph.D. degree with the Omni Lab for Visual Engineering and Science (OLIVES). She is working in the fields of machine learning, deep learning, computer vision, and image processing. Her research interests include the robustness of machine learning models and learning with limited data.
\end{IEEEbiography}

\begin{IEEEbiography}[{\includegraphics[width=1in,height=1.25in,clip,keepaspectratio]{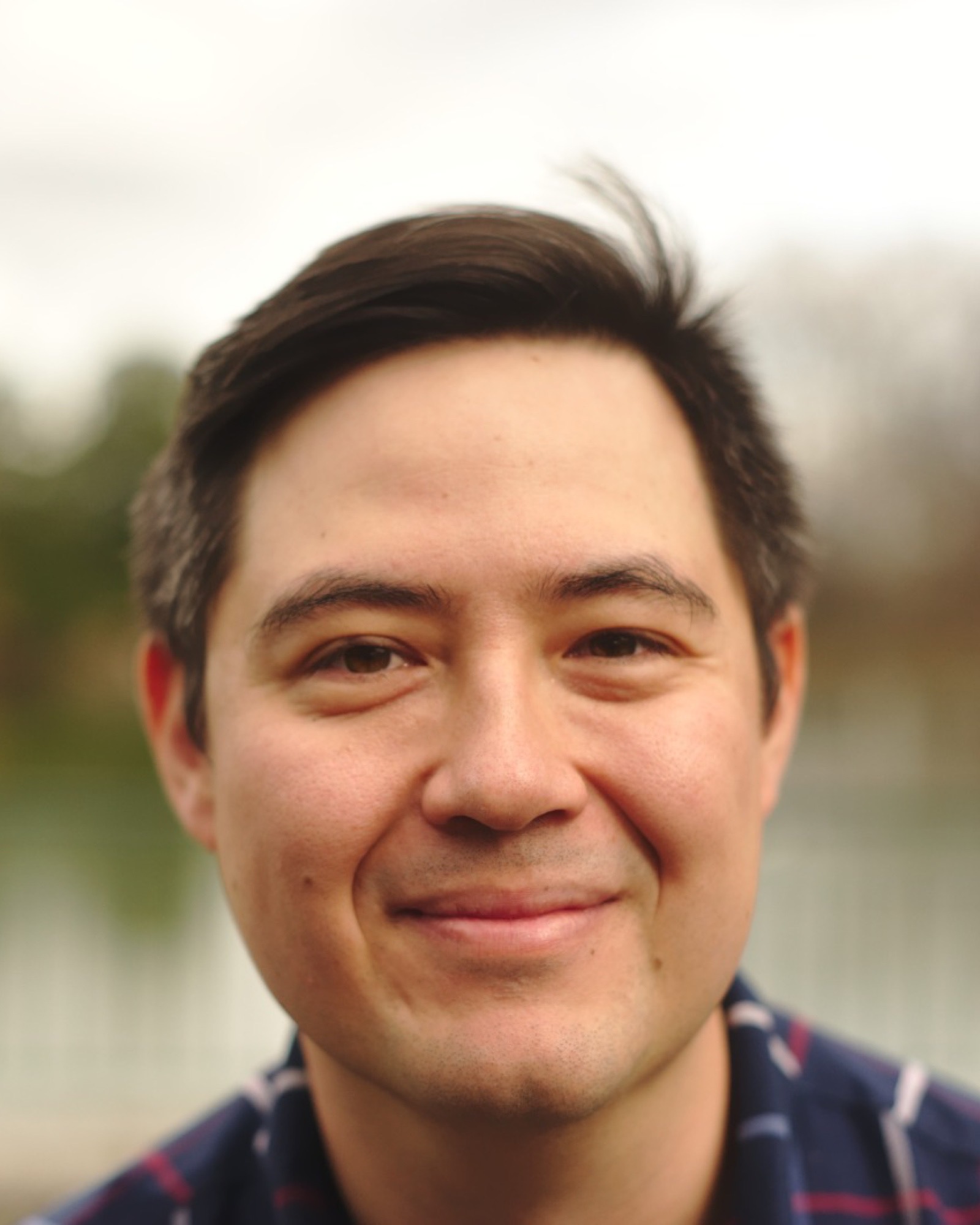}}]{Charlie Lehman} is a Machine Learning Ph.D. student at the Georgia Tech Omni Lab for Visual Engineering and Science (OLIVES). His research interests include robustness and explainability of deep vision models. He is also an Engineering Duty Officer with the U.S. Navy Reserves, where he specializes in the maintenance and repair of surface vessels.
\end{IEEEbiography}

\begin{IEEEbiography}[{\includegraphics[width=1in,height=1.25in,clip,keepaspectratio]{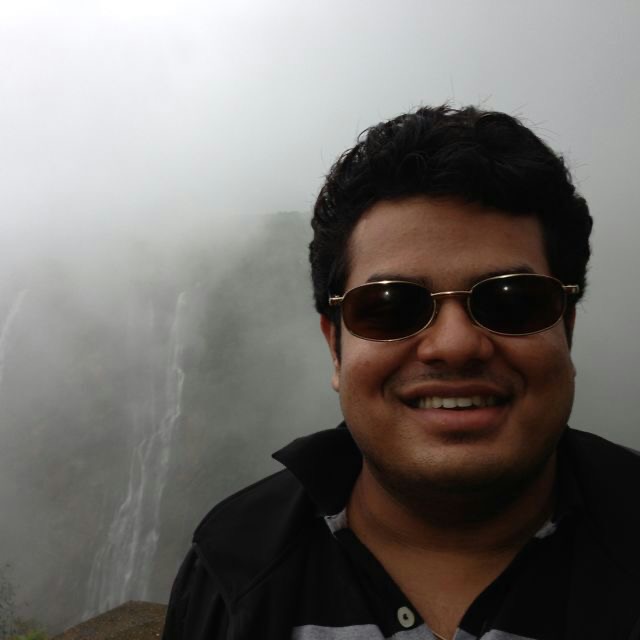}}]{Mohit Prabhushankar} (Member, IEEE) received his Ph.D. degree in electrical engineering from the Georgia Institute of Technology in 2021. He is currently a Postdoctoral Research Fellow in the School of Electrical and Computer Engineering at the Georgia Institute of Technology in the Omni Lab for Intelligent Visual Engineering and Science (OLIVES). He is working in the fields of image processing, machine learning, active learning, healthcare, and robust and explainable AI. He is the recipient of the Best Paper award at ICIP 2019 and Top Viewed Special Session Paper Award at ICIP 2020. He is the recipient of the ECE Outstanding Graduate Teaching Award, the CSIP Research award, and of the Roger P Webb ECE Graduate Research Assistant Excellence award, all in 2022.
\end{IEEEbiography}

\begin{IEEEbiography}[{\includegraphics[width=1in,height=1.25in,clip,keepaspectratio]{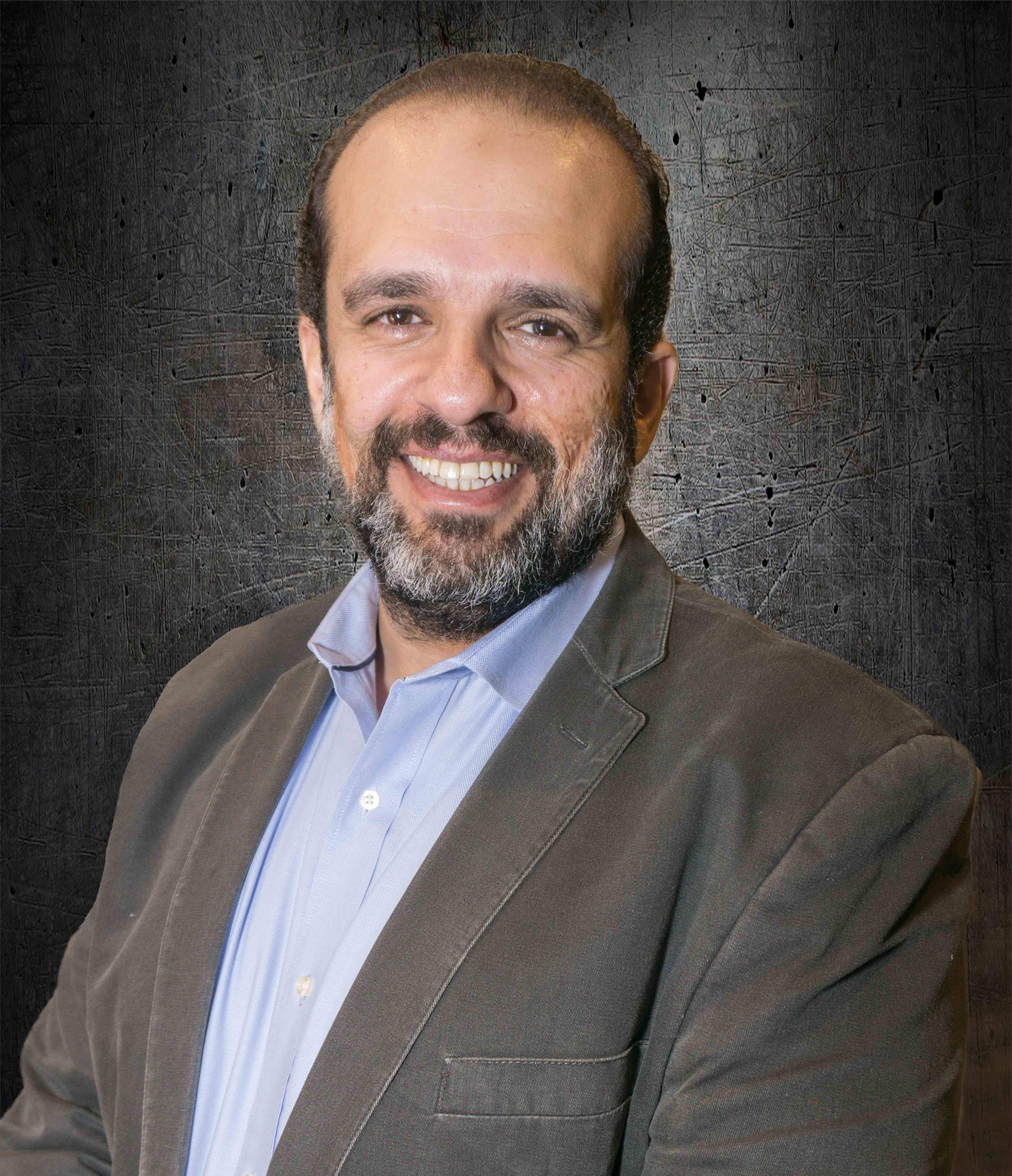}}]{Ghassan AlRegib} (Fellow, IEEE) is currently the John and Marilu McCarty Chair Professor in the School of Electrical and Computer Engineering at the Georgia Institute of Technology. In the Omni Lab for Intelligent Visual Engineering and Science (OLIVES), he and his group work on robust and interpretable machine learning algorithms, uncertainty and trust, and human-in-the-loop algorithms. The group has demonstrated their work on a wide range of applications such as Autonomous Systems, Medical Imaging, and Subsurface Imaging. The group is interested in advancing the fundamentals as well as the deployment of such systems in real-world scenarios. He has been issued several U.S. patents and invention disclosures. He is active in the IEEE as an IEEE Fellow. He served on the editorial board of several transactions and served as the TPC Chair for ICIP 2020, ICIP 2024, and GlobalSIP 2014. He was area editor for the IEEE Signal Processing Magazine. In 2008, he received the ECE Outstanding Junior Faculty Member Award. In 2017, he received the 2017 Denning Faculty Award for Global Engagement. He and his students received the Best Paper Award in ICIP 2019. 

\end{IEEEbiography}

\EOD

\end{document}